\documentclass{article}

\usepackage{arxiv}

\usepackage[utf8]{inputenc} 
\usepackage[T1]{fontenc}    

\usepackage[bookmarks=false]{hyperref}
    \hypersetup{colorlinks,
      linkcolor=blue,
      citecolor=blue,
      urlcolor=blue}
      
\usepackage{url}            
\usepackage{booktabs}       
\usepackage{amsfonts}       
\usepackage{nicefrac}       
\usepackage{microtype}      
\usepackage{lipsum}		
\usepackage{graphicx}
\usepackage{doi}

\usepackage{amsmath}
\usepackage{amssymb}

\RequirePackage{comment}

\usepackage{siunitx}
\DeclareSIUnit[number-unit-product={}]{\percent}{\%}

\usepackage[numbers]{natbib} 

\usepackage{xspace}

\usepackage[figuresright]{rotating}

\RequirePackage[english, textsize=footnotesize, backgroundcolor=blue!20!white, colorinlistoftodos,prependcaption, bordercolor=red]{todonotes} 

\newcommand{\robotskill}{robot skill\xspace}
\newcommand{\robotskills}{robot skills\xspace}
\newcommand{\RobotSkills}{Robot Skills\xspace}
\newcommand{\pih}{peg-in-hole\xspace}
\newcommand{\elta}{electrical terminal assembly\xspace}
\newcommand{\ELTA}{Electrical Terminal Assembly\xspace}
\newcommand{\snaphook}{snap-hook\xspace}
\newcommand{\snaphooks}{snap-hooks\xspace}
\newcommand{\SnapHook}{Snap-Hook\xspace}
\newcommand{\DINrail}{DIN-rail\xspace}

\newcommand{\SimtoReal}{Sim-to-Real\xspace}

\newcommand{\figref}[1]{Fig.~\ref{#1}}   
\newcommand{\tabref}[1]{Table~\ref{#1}}  

\DeclareMathOperator*{\argmax}{arg\,max}
\DeclareMathOperator{\atantwo}{atan2}  

\usepackage{mathtools}
\newcommand*{\coloneqq}{\mathrel{\vcenter{\baselineskip0.5ex \lineskiplimit0pt
                     \hbox{\scriptsize.}\hbox{\scriptsize.}}}=}
\newcommand*{\rcoloneqq}{=%
                     \mathrel{\vcenter{\baselineskip0.5ex \lineskiplimit0pt
                     \hbox{\scriptsize.}\hbox{\scriptsize.}}}}

\begin{document}



\title{Simulation-based Learning of Electrical Cabinet Assembly Using \RobotSkills}

\date{February 16, 2026}	


\author{ Arik Lämmle\thanks{Corresponding author. {\it E-mail address:} arik.laemmle@cellios.de | arik.laemmle@ipa-extern.fraunhofer.de} \\
	formerly Fraunhofer Institute for Manufacturing\\
    Engineering and Automation IPA\\
	Stuttgart, Germany \\
    \And
	Philipp Tenbrock \\
	formerly Fraunhofer Institute for Manufacturing\\
    Engineering and Automation IPA\\
	Stuttgart, Germany \\
	\And
	Balázs András Bálint\\
	formerly Fraunhofer Institute for Manufacturing\\
    Engineering and Automation IPA\\
	Stuttgart, Germany\\
	\And
	David Traunecker \\
	formerly Institut für Strahlwerkzeuge IFSW\\
    University of Stuttgart\\
    Stuttgart, Germany
	\And
	Frank Nägele\\
	formerly Fraunhofer Institute for Manufacturing\\
    Engineering and Automation IPA\\
	Stuttgart, Germany\\
	\And
	József Váncza\\
	Research Laboratory on Engineering and \\Management Intelligence, Institute for \\Computer Science and Control SZTAKI\\
    Budapest, Hungary
	\And
	Marco F. Huber\\
    Institute of Industrial Manufacturing and \\Management IFF, University of Stuttgart,\\
	Fraunhofer Institute for Manufacturing\\
    Engineering and Automation IPA\\
	Stuttgart, Germany
}


\renewcommand{\undertitle}{}
\renewcommand{\shorttitle}{}

\hypersetup{
pdftitle={Simulation-based Learning of Electrical Cabinet Assembly Using Robot Skills},
pdfsubject={Simulation-based Learning of Electrical Cabinet Assembly Using Robot Skills}, 
pdfauthor={Arik Lämmle},
pdfkeywords={Assembly Automation; Deep Reinforcement Learning; Physics Simulation},
}


\maketitle

\begin{abstract}
This paper presents a simulation-driven approach for automating the force-controlled assembly of electrical terminals on DIN-rails, a task traditionally hindered by high programming effort and product variability. The proposed method integrates deep reinforcement learning (DRL) with parameterizable robot skills in a physics-based simulation environment. To realistically model the snap-fit assembly process, we develop and evaluate two types of joining models: analytical models based on beam theory and rigid-body models implemented in the MuJoCo physics engine. These models enable accurate simulation of interaction forces, essential for training DRL agents. The robot skills are structured using the pitasc framework, allowing modular, reusable control strategies.
Training is conducted in simulation using Soft Actor-Critic (SAC) and Twin Delayed Deep Deterministic Policy Gradient (TD3) algorithms. Domain randomization is applied to improve robustness. The trained policies are transferred to a physical UR10e robot system without additional tuning.
Experimental results demonstrate high success rates (up to 100\%) in both simulation and real-world settings, even under significant positional and rotational deviations. The system generalizes well to new terminal types and positions, significantly reducing manual programming effort.
This work highlights the potential of combining simulation-based learning with modular robot skills for flexible, scalable automation in small-batch manufacturing. Future work will explore hybrid learning methods, automated environment parameterization, and further refinement of joining models for design integration.
\end{abstract}

\section{Introduction}\label{sec:intro}

\begin{figure*}[!ht]
    \vspace*{8pt}
    \centering
    \includegraphics[width=\textwidth]{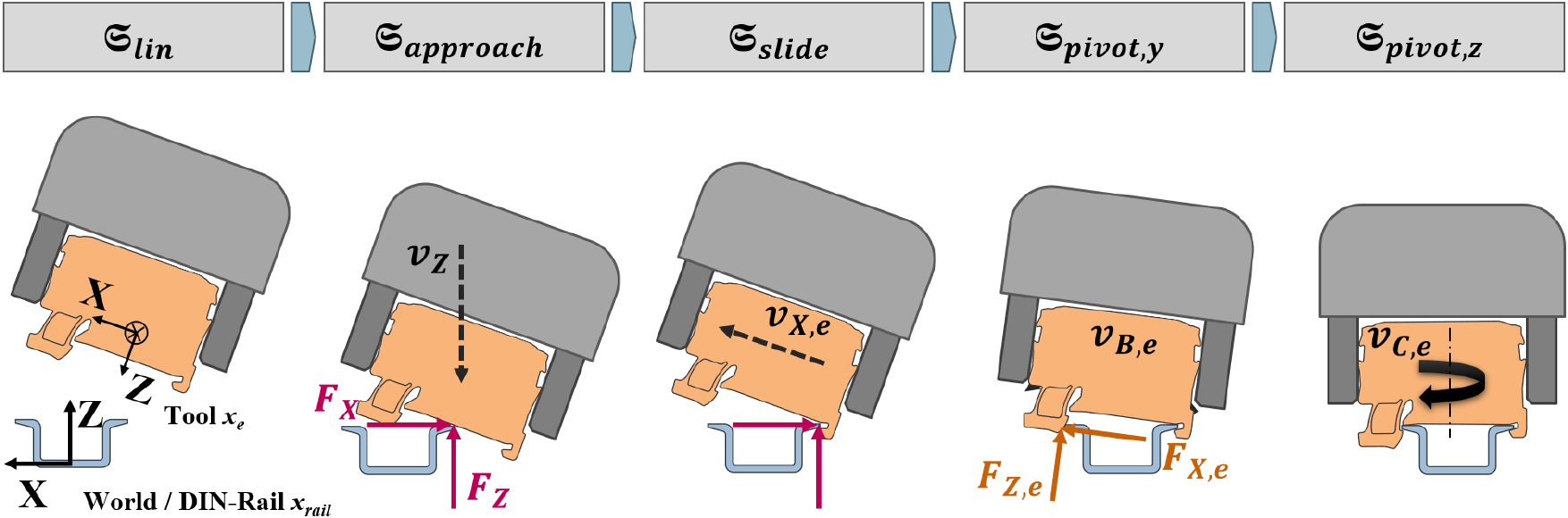}
    \caption{Structure of the \elta task, used in the presented work.}
    \label{fig:2_Eletrical-Terminal-Assembly_Process_Overview}
    \vspace*{-8pt}
\end{figure*}

Robot-based automation of force-controlled assembly tasks is still challenging and time-consuming, even when using advanced or intuitive robot programming tools. Decreasing batch sizes and a high diversity of variants of the products to be assembled, not least as a result of growing product personalisation, additionally require high-frequency adaptations of the robot program. Thus, in many cases, the effort required for programming and setting up the robotic system exceeds the time required for manual product assembly. Many assembly tasks show high potential for automation, however, most of these remain unexploited due to the high effort required for robot programming, a lack of flexibility in changeover, and the ability to automatically adjust to uncertainties and tolerances on the robot’s part.

The assembly of electrical terminals in control cabinet manufacturing is a prominent example. Although the underlying assembly process can be automated well with established technologies, it is rarely realised in industrial practice due to the high number of variants and the associated programming effort for the robot to adapt to these variants \cite{Tempel.2017}.

One possible solution for assembly automation capable of such adaptation is to train robot controllers on a real-world system or inside a simulation using \emph{machine learning} (ML) \cite{Ibarz.2021}. Both approaches rely on training data that accurately represent the assembly process. Using a simulation allows applying additional variation to the training domain, such as domain randomization of process characteristics, to increase the trained controllers’ capability to generalize and handle uncertainty \cite{Wang.2026}.

Another solution to reduce the programming effort is the application of predefined modules or \robotskills. These \robotskills comprise ready-to-use position- and force-controlled programme modules that can be adapted to the products to be assembled or the processes to be carried out by setting parameters. Variants, for example, in the assembly of comparable electrical terminals, can be easily realised by setting these parameters. Each skill and program module encapsulates the relationship between the robot's movement commands and the sensor observations. At the same time, despite the advantages of \robotskills, the skill parameters still have to be defined manually by an expert, limiting their application, especially for small batch sizes.

The work presented in this paper combines the strengths of both offline learning from a physics simulation environment with the advantages of predefined, position- and force-controlled \robotskills. We propose the adaptation of a simulation-based task-level \emph{deep reinforcement learning} (DRL) pipeline of robot control policies using \robotskills \cite{Laemmle.2020c} to the task of mounting electrical cabinet terminals on DIN-rails. \figref{fig:2_Eletrical-Terminal-Assembly_Process_Overview} illustrates the use case. Thus, we ultimately aim to reduce the time required for skill parameterisation by transferring the task to an intelligent, learning-based entity. Furthermore, we already introduce process- and product-specific uncertainties and tolerances during training in the simulation, allowing the robot agent to learn the situation-based adaptation to industry-typical deviations.

The assembly of electrical terminals on DIN-rails is based on deformable \snaphooks to realize the mechanical connection between both components.
However, most established state-of-the-art rigid multi-body simulators for training robotic agents do not provide sufficient functionalities for calculating deformation forces.
Thus, we initially determine the core problem to digitally represent the real-world scenario of electrical terminal assembly and enable in silico training data generation for applying DRL. We examine the capabilities of the (widely accepted) \emph{physics engine} (PE) MuJoCo \cite{Todorov.2012} and establish two approaches as possible solutions. The first approach of rigid body joining models solely uses existing features of the PE without extending its software. The second approach, the analytical joining models, is instead a custom extension to the PE that aims to be a computationally more efficient alternative.

Subsequently, we train predefined \robotskills in the previously extended simulation environment. Two off-policy learning algorithms, Soft-Actor-Critic (SAC) and Twin Delayed Deep Deterministic Policy Gradient (TD3), are used to train the agent. Overall, we demonstrate the capabilities of the trained robot agent to handle industry-typical position and orientation tolerances by adapting the \robotskills depending on the specific situation. We demonstrate the effectiveness of training \robotskills both in simulation and on a physical robot.

The paper is structured as follows. Section \ref{sec:related-research} describes the considered problems of simulating the \elta and skill parameterization as well as related research. Section \ref{sec:system-design} provides details of the system design and the skill formalism. The developed joining models for force approximation are presented in Section \ref{sec:joining-models}. Section \ref{sec:learning} focuses on simulation-based training while Section \ref{sec:evaluation} presents results from the subsequent transfer to a physical robot system and real-world experiments. Finally, we summarize and critically assess the conducted research and provide an outlook on future work.

The interested reader is referred to our online video recordings of the training and the execution of the \elta on the physical robot system: \url{https://www.youtube.com/watch?v=7tAaDgHLIts}.

\section{Related Research and Considered Problem}
\label{sec:related-research}

In the following, related research and development activities will be discussed. In addition, the robot-based process of the \elta and the problem considered in the presented work are described. To solve the problem, we employ a combination of force-controlled, hardware-independent robot kills, a physics simulation extended specifically for modeling \snaphooks, and established deep reinforcement learning methods.

\subsection{Force-controlled \RobotSkills}

Continuing previous work~\cite{Laemmle.2022c}, we employ the pitasc \robotskill framework~\cite{Nagele.2018, Nagele.2021} to create our robot program. Skills define a layer of abstraction over the robot's control and allow program creation in the task domain enabling programming from a process expert point of view~\cite{Nagele.2019}. A skill-based programming approach also articulates and constrains the parameter domain of the program, making the optimization problem more tractable and a solution faster to develop automatically~\cite{Kroemer.2019}. The pitasc framework facilitates reusability even across manipulators through its hierarchical structure that abstracts the properties of kinematics at low hierarchical levels.

\subsection{Simulation of \SnapHook Assembly}

Snap-fits or \snaphooks ~\cite{Tempel.2017} are among the most preferred mechanical joinings for polymer parts in industrial assemblies~\cite{Troughton.2008} today, but the robotics community does not discuss their assembly task proportionately. Nonetheless, analytical models exist~\cite{Kunz.2000, Amaya.2019, Kakade.2020}, providing force estimates during the joining assembly. However, they usually use the components' deformation state as input, which is generally unavailable in today's widely adopted physics simulation tools as they primarily handle rigid multi-body systems.

While finite-element simulations have become a standard part of modern CAD software for stress analysis simulation~\cite{FEM.solidworks, FEM.inventor, FEM.creo}, the feature is mainly geared towards design validation, trading accuracy for a slower-than-real-time running speed~\cite{Rossdeutscher.2011}. Also, the results are generally not accessible by third-party software, prohibiting the feature's integration with, e.g., machine learning pipelines entirely.

Using a real-world process as a data source for the task's robotic learning is also hindered. While \snaphooks can be designed for repeated assembly and disassembly, most designs specifically hinder the separation of the connected parts, making the task non-reversible and developing a resettable digital replica even more desirable.

Establishing a digital model of soft bodies and their deformation remains the core problem that one has to solve to digitally represent the \snaphook assembly and any industrial use case that incorporates it, such as the \elta. For off-the-shelf dynamic simulators or physics engines, e.g., MuJoCo, such a model would transform kinematic data to forces and torques that can be applied in the virtual environment.

\subsection{Learning Robot-based Assembly}

Training robot controllers using ML, especially in the case of DRL, is a rising topic in robotics to ease the programming of even complex assembly tasks~\cite{ElShamouty.2019, Zhu.2026} by alleviating the need for problem-specific professional knowledge and limiting the need for complicated and repetitive reprogramming, as such controllers can potentially adapt to previously unseen scenarios~\cite{Laemmle.2020c,Manyar.2026}. However, DRL has the disadvantage of requiring enormous amounts of data, so training a robot with real hardware is expensive in terms of money and time, especially for non-reversible tasks, i.e., tasks whose completion results in changes that cannot be reversed trivially or without human intervention. Thus, the advantages of generating data from a simulation are apparent: relatively low costs, vast amounts of data easily accessible, and potentially faster-than-real-time execution~\cite{Ibarz.2021}. However, due to the sim-to-real gap, policies trained in simulation can perform poorly in reality~\cite{Ibarz.2021}, with one of the reasons being unmodeled dynamics~\cite{Tan.2018}. Therefore, we choose the physics engine MuJoCo~\cite{Todorov.2012} to build our virtual environments in this work for the engine's accuracy in dynamics simulation and wide adoption by the robotics community. Moreover, force-based robot control relies on the interaction forces between the robot and its environment; therefore, we focus on generating accurate synthetic interaction forces between the electrical terminals and the \DINrail in the electrical cabinet assembly to reduce the sim-to-real gap.\\ 

Nonetheless, similar research has been carried out on training robot agents to perform assembly tasks both in simulation and the physical domain. While the \pih task is a well-researched benchmark process ~\cite{Inoue.2017,Fan.2019,park.2020} for robot-based and force-controlled assembly, the \elta has been little researched in the past.  



\subsection{Robot-based \ELTA}

According to N{\"a}gele ~\cite{Nagele.2019}, a skill is denoted by $\mathfrak{S}$, while $\mathfrak{S}_{i}$ indicates a specific skill within a finite set of skills. As depicted in \figref{fig:2_Eletrical-Terminal-Assembly_Process_Overview} and described in \eqref{eq:skill-def-ETA}, the robot-based assembly of electrical terminals on a \DINrail typically involves three successive movements $\mathfrak{S}_{approach}$, $\mathfrak{S}_{slide}$ and $\mathfrak{S}_{pivot}$, accompanied by two auxiliary movements $\mathfrak{S}_{lin}$ at the beginning and $\mathfrak{S}_{pivot}$ at the end of the process ~\cite{Nagele.2019}. During the first auxiliary movement $\mathfrak{S}_{lin}$, the robot proceeds to a defined pre-position. Following this, we assume the terminal is already in the gripper mounted on the robot. Subsequently, the robot moves the gripped terminal in the $\mathfrak{S}_{approach}$ skill linearly along the Z-axis of the world coordinate system towards the \DINrail. Once the robot detects a contact force between the terminal and the \DINrail, the $\mathfrak{S}_{slide}$ skill is activated. The robot applies a constant contact force to the \DINrail and traverses linearly along the X-axis of the terminal. The skill is completed once the robot establishes contact between the fixed hook on the terminal and the right side of the \DINrail. During the subsequent $\mathfrak{S}_{pivot}$ skill, the robot rotates the terminal around the previously identified contact point between the \DINrail and the fixed hook. During the rotational movement, the deformable \snaphook of the terminal engages with the left side of the \DINrail and is deflected as the rotational movement continues. As a result of the deflection and deformation, the \snaphook exerts assembly forces on the robot. The assembly force along the X-axis of the terminal is referred to as the lateral force $F_{Q}$ and the force along the Z-axis of the terminal as the joining force $F_{J}$. After exceeding the complete deflection of the \snaphook, a form-fitting and partly also force-fitting coupling is established between the terminal and the \DINrail. At the completion of the \elta, the robot can perform another $\mathfrak{S}_{pivot}$ skill around the vertical axis of the terminal to adjust its orientation on the \DINrail, subject to the terminal design being suitable for this additional movement.\\

So far, to the best of our knowledge, no solution exists for the simulation-based learning of the \elta, employing position- and force-controlled skills, beside the work presented in this paper and the research by Monnet et al. \cite{Monnet.2025}. At the same time, current developments address one partial aspect of the work presented.

Lin et al. ~\cite{Lin.2024} train their agent to execute insertion tasks using TD3 learning algorithm based on manual task demonstrations. The authors reach success rates of up to $90 \: \%$, incorporating position uncertainties of up to $2 \: mm$.
Chen et al. ~\cite{Chen.2024} also incorporate uncertainties in their process execution. They achieve success rates up to $90 \: \%$ for the \pih task in simulation and between $63.3 \: \%$ and $83.3 \: \%$ for executions on the physical robot. 
Training a robotic agent in simulation is also done by Zhang et al. ~\cite{Zhang.2023b}. While the authors avoid explorations and assume the target pose to be known a priori, they achieve success rates between $70 \: \%$ to $90 \: \%$ even for tight part clearances down to $0.02 \: mm$.
Additional processes trained in simulation besides the \pih task are researched by Tang et al. ~\cite{Tang.2023}. While the authors solve the \pih task with a success rate of  $76,7 \: \%$, the more challenging gear and connector assembly are even successful in $92,5 \: \%$ and $85,0 \: \%$, respectively.

\section{System Design}
\label{sec:system-design}

\begin{figure*}[!ht]
    \centering
    \includegraphics[width=\textwidth]{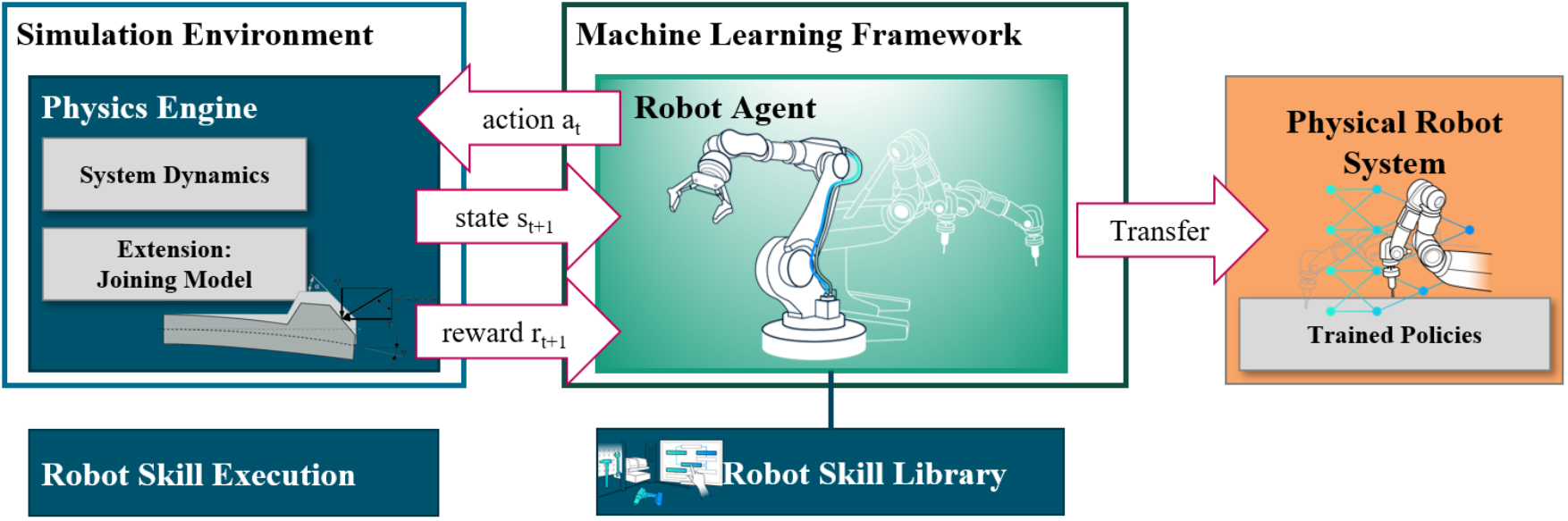}
    \caption{Proposed framework for learning \robotskill parameters from simulation.}
    \label{fig:learning_framework}
    \vspace*{-8pt}
\end{figure*}

For the simulation-based training of \robotskills, we propose the learning framework presented in \figref{fig:learning_framework} comprising three core components. The \textbf{i) physics simulation environment} approximates the system dynamics $p(s_{t+1}|s_{t}, a_{t})$ of the real world and thus, the interaction of the robot agent with its environment. An action $a_t$ of the agent in state $s_t$ at time $t$ transfers the system into the subsequent state $s_{t+1}$. The situational selection of the following target-oriented action $a_{t+1}$ by the agent is the goal of the actual learning and thus, of the \textbf{ii) machine learning framework}. Possible actions of the robot agent include the selective choice of suitable \robotskills and parameterizing the skills accordingly. During the training in the simulation, the robot agent is provided with the same observation capabilities via sensors as on the physical system. Finally, the learning framework incorporates \textbf{iii) transfer mechanisms} to migrate the trained robot control algorithms from the source domain (simulation) to the target domain (real world).

\subsection{Physics Simulation}

In the presented work, we employ the physical simulation environment MuJoCo, widely adopted in the machine learning community, to simulate the robots interaction with its environment. The general equation of motion of the robot in continuous time is \begin{equation}\label{eq:Dynamic-Model-Robot}
    M(q)\dot{v}_{e}+C(q,\dot{q})v_{e}+g(q)=h_{c}-h_{e},
\end{equation}
where $M \in \mathbb{R}^{6\times6}$ denotes the operational space inertia matrix, $C \in \mathbb{R}^{6\times1}$ describes the centrifugal and Coriolis effects, and $g \in \mathbb{R}^{6\times1}$ represents the gravitational effects. $q \in \mathbb{R}^{6\times1}$ denotes the vector of the joint coordinates of the 6-axis robot. 
The joint coordinates $q$ are related to the the position $p_{e}$ and the orientation $\varphi_{e}$ of the robot end-effector or any coordinate system with respect to a fixed base frame, by means of the Jacobian $J \in \mathbb{R}^{6\times6}$

\begin{equation}\label{eq:end-effector-pose}
    x_{e} = J q = \left( p_{e},\varphi_{e}\right) \in \mathbb{R}^{6\times1}.
\end{equation}

On the right hand side of the equation of motion \eqref{eq:Dynamic-Model-Robot}, the vector $h_{c} \in \mathbb{R}^{6\times1}$ of the active torques at the joints of the robot and the wrench 
\begin{equation}
    h_{e} = \left( F_{e}, M_{e}\right) \in \mathbb{R}^{6\times1}
\end{equation}
comprising the external acting forces $F_{e} \in \mathbb{R}^{3\times1}$ and torques $M_{e} \in \mathbb{R}^{3\times1}$ related to the end-effector coordinate system of the robot are given.

Being a rigid body simulation at heart, MuJoCo is well suited for approximating the external forces and moments acting during the execution of the first three \robotskills $\mathfrak{S}_{lin}$, $\mathfrak{S}_{approach}$ and $\mathfrak{S}_{slide}$. At the same time, the accuracy of the simulation strongly depends on the chosen simulation parameters. In order to obtain an accurate approximation of the robot's interaction with its environment, the simulation parameters were optimised manually. Experimental investigations of the \elta on a physical robot system served as a basis for comparison.
During the execution of the two pivot skills $\mathfrak{S}_{pivot}$, the \snaphook is deformed and exerts forces on the robot depending on its deflection, the geometry of the \snaphook and the material properties of the terminal. Consequently, additional joining models have to extend the rigid body simulation MuJoCo to approximate the acting forces during the \snaphook assembly. In Section \ref{sec:joining-models}, two different model classes are developed and integrated into the simulation environment.

\subsection{Skill Formalism for the \ELTA}

For the research presented, we employ the skill model of \cite{Nagele.2018,Nagele.2021}. Each skill 
\begin{equation}\label{eq:skill-def}
    \mathfrak{S}=(\mathcal{N,KE,T,SC,M,TR},\mathfrak{S}_{sub})
\end{equation}
is described as a 7-tuple with the unique name $\mathcal{N}$ of the skill, a set of kinematic elements $\mathcal{KE}$ describing the kinematics of the robot and the application to be performed based on the iTasC formalism \cite{DeSchutter.2007} and a list of tasks $\mathcal{T}$ as well as the corresponding control variables and their parameters. In addition, $\mathcal{SC}$ contains additional scripts and support functions, such as opening and closing a gripper. The monitors $\mathcal{M}$ contain stop conditions for terminating the skill, whereas $\mathcal{TR}$ includes transition conditions for the skill itself and its set of subordinate skills $\mathfrak{S}_{sub}$.

The high-level skill for the \elta 
\begin{equation}\label{eq:skill-def-ETA}
    \mathfrak{S}_{terminal} = (\mathfrak{S}_{lin},\mathfrak{S}_{approach}, \mathfrak{S}_{slide}, \mathfrak{S}_{pivot}, \mathfrak{S}_{pivot})
\end{equation}
comprises five subordinate skills. To limit the number of attempts, the agent may use up to N = 6 consecutive skill executions 
\begin{equation}
    \mathfrak{S}_{overall} = (\mathfrak{S}_{i,1}, ... ,\mathfrak{S}_{i,N}) \: \: with \: \: \mathfrak{S}_{i} = \mathfrak{S}_{terminal} \vee \mathfrak{S}_{pivot}
\end{equation}
of the whole skill sequence $\mathfrak{S}_{terminal}$ or individual pivot skills $\mathfrak{S}_{pivot}$ available for assembling a terminal. Early training runs indicated that this provides the agent with a sufficient number of actions to learn how to compensate for inaccuracies and tolerances as well as to explore its environment.
During training and subsequent execution on the real system, the robot agent can choose the parameters $\mathcal{T}_{i,t}$ of the individual skills depending on the current situation observed through its sensors.


\subsection{Reinforcement Learning Framework}
\label{subsec:RL-Framework}

The training aims at finding a policy for solving the mixed-integer optimisation problem of skill selection and parameterisation. Accordingly, the reinforcement learning objective is to obtain an optimal policy 
\begin{equation}\label{eq:RL-objective}
    \begin{aligned}
        \pi^* & = \argmax_{\pi} G(\pi) \nonumber\\
              & = \argmax_{\pi} \mathbb{E}_{\tau \sim p(\tau|\pi)} \left[\sum_{t=1}^{\textrm{T}} \gamma^{t-1} r(o_t,a_t)\right],
    \end{aligned}
\end{equation}
that, depending on the current state $s_t$, selects the subsequent action $a_t$ of the robot agent, as defined in the following, to maximise the expected discounted return. The sequence of states and chosen actions is called the T-step trajectory $\tau$ with probability distribution $p(\tau|\pi)$.

Utilising its sensors, the robot agent can access the observation 
\begin{equation}\label{eq:system-model-observation}
    o_{t} = \left[p_{rel,t},q_{t}, h_{e,t} \right] \in \mathbb{R}^{13}
\end{equation}
at time $t$ as a subset of the fully observable state $o_t \subset s_t$. This observation is defined in state space form containing the current relative position $p_{rel,t} \in \mathbb{R}^{3\times1}$ and $q_{t} \in \mathbb{R}^{4\times1}$ the quaternions describing the relative orientation between the reference coordinate system of the \DINrail and the tool coordinate system of the terminal as well as the forces and moments $h_{e} = (F_{e}, M_{e}) \in \mathbb{R}^{6\times1}$ acting on the robot measured in the coordinate system of the end effector.

Depending on the individual skill, various learnable action configurations $\hat{\mathcal{T}}_{i}$ are available to the robot agent during training. In addition to the skill parameters listed below, the agent chooses between the skill $\mathfrak{S}_{terminal}$ or $\mathfrak{S}_{pivot}$ to be executed in a higher-level action. For a better understanding of the individual skill parameters and actions, please refer to their graphical representation in \figref{fig:2_Eletrical-Terminal-Assembly_Process_Overview}.

In the following, the learnable parameters of the skill are introduced. During the first skill 
\begin{equation}
    \hat{\mathcal{T}}_{lin} = (\Delta p_{X,lin},\varphi_{B,lin}),
\end{equation}
 the agent can change the pre-position $\Delta p_{X,lin} \in [-25 \: \textrm{mm}, 25\: \textrm{mm}]$ of the gripped terminal and its orientation $\varphi_{B,lin} \in [-5 \: \textrm{°},5 \: \textrm{°}]$. 

For the subsequent approach movement 
\begin{equation}
    \hat{\mathcal{T}}_{approach} = (v_{Z,approach},F _{Z,approach}),
\end{equation}
the agent trains both the linear movement speed $v_{Z,approach} \in [2 \: \frac{\textrm{mm}}{\textrm{s}}, 20 \: \frac{ \textrm{mm}}{\textrm{s}}]$ and the contact force $F _{Z,approach} \in [3.0 \: \textrm{N}, 15.0 \: \textrm{N}]$ to be applied. 

The third skill 
\begin{equation}
    \hat{\mathcal{T}}_{slide} = (v_{XZ,slide},F _{slide},F _{XZ,slide},C_{PD,slide})
\end{equation}
is defined by the learnable movement speed $v_{XZ,slide} \in [1 \: \frac{\textrm{mm}}{\textrm{s}}, 10 \: \frac{\textrm{mm}}{\textrm{s}}]$, the applied contact force $F _{slide} \in [1.0 \: \textrm{N}, 30.0 \: \textrm{N}]$, the desired target force $F _{slide,target} \in [1.0 \:\textrm{N}, 15.0 \: \textrm{N}]$ and the compliance of the used controller $C_{PD,slide} \in [0.0001 \frac{\textrm{Nm}}{\textrm{s}}, 0.001 \frac{\textrm{Nm}}{\textrm{s}}]$. 

Finally, the agent parameterises the pivot skill 
\begin{equation}
    \hat{\mathcal{T}}_{pivot} = (\varphi_{B,pivot},\omega_{AB,pivot},\varphi_{C,pivot},F _{X,pivot},F _{Z,pivot})
\end{equation}
by setting the rotation angle $\varphi_{B,pivot} \in [-30 \: \textrm{°},30 \: \textrm{°}]$, the rotation speed $\omega_{AB,pivot} \in [0.02 \: \frac{\textrm{rad}}{\textrm{s}}, 0.5 \: \frac{\textrm{rad}}{\textrm{s}}]$, and the forces $F _{X,pivot} \in [3.0 \: \textrm{N}, 30.0 \: \textrm{N}]$ and $F _{Z,pivot} \in [3.0 \: \textrm{N}, 30.0 \: \textrm{N}]$ to be applied. By additionally selecting the rotation angle $\varphi_{C,pivot} \in [-20 \: \textrm{°},20 \: \textrm{°}]$ around the vertical axis of the terminal, the agent can both correct rotations of the terminal on the \DINrail or compensate for inaccuracies in the orientation of the \DINrail itself. In general, the robot agent is intended to deal with inaccuracies in the process and tolerances of the components to be assembled by introducing them into the simulation during training. For this purpose, at the beginning of a training run, the position of the \DINrail along the X-axis is uniformly randomised in the interval $[-5 \: \textrm{mm},5\: \textrm{mm}]$ and the orientation in the interval $[-3 \: \textrm{°},3\: \textrm{°}]$. Furthermore, deviations of the force-moment sensor used are modelled by randomising the measured forces from the normal distribution $\mathcal{N}\left(0\:\textrm{N}, 0.2^2\:\textrm{N}^2\right)$.

Resulting from a chosen and executed action $a_t$, the system state is transferred from $s_t$ to $s_{t+1}$. In addition, during training, the agent receives the distance-based reward 
\begin{equation} \label{eq:eta-train-reward}
     r(s_t,a_t) = - \frac{\lvert d_t \rvert}{\lvert d_{norm} \rvert},  
\end{equation}
where $d_t$ is a measure of the relative distance between the terminal and its target position and orientation on the \DINrail. At the same time, $d_{norm}$ is an arbitrarily chosen distance to define the reward at $r = -1$.
\section{Joining Models for \SnapHook Simulation} 
\label{sec:joining-models}

Training force-controlled \robotskills demands a realistic simulation of the system dynamics and the forces acting on the robot agent during the \elta. Therefore, MuJoCo is extended by additional joining models since the default simulation of rigid bodies simulation is not suitable for the force approximation of deformable \snaphooks on terminals. These joining models approximate the acting, external assembly forces $h_{e,t}$ as a parametrisable function 
\begin{equation}\label{eq:goal-joining-model}
    f_{\theta} (\ensuremath{P}_{product},s_{t})=h_{e,t}
\end{equation}
of the product properties $\ensuremath{P}_{product}$ of the terminal and the \snaphook as well as selective aspects of the current system state $s_t$, e.g., the present relative position between the terminal and the \DINrail. Two different solutions are proposed for modelling the joining models and identifying product- and process-specific model parameters $\theta$: \textbf{1)} external analytical joining models approximating the acting forces and providing them back to MuJoCo as output variables and \textbf{2)} custom rigid-body joining models employing the available functional capacities in MuJoCo.\\

During the assembly of snap-hook-based components such as electrical terminals, the \snaphook, and its bending beam are deformed and deflected by a fixed snap-in locking fixture. As a result of the deformation, the lateral force $F_Q$ acts perpendicularly, and the joining force $F_J$ acts parallel to the direction of joining. After the \snaphook has been fully deflected, the bending beam rebounds, resulting in a form-fitting connection. In some cases, a residual deflection and deformation of the bending beam remains, resulting in an additional force-fitting connection.

\subsection{Analytical Joining Models}\label{subsec:model-analytical}

\begin{figure}[!tb]
    \vspace*{8pt}
    \centering
    \includegraphics[width=3.5in]{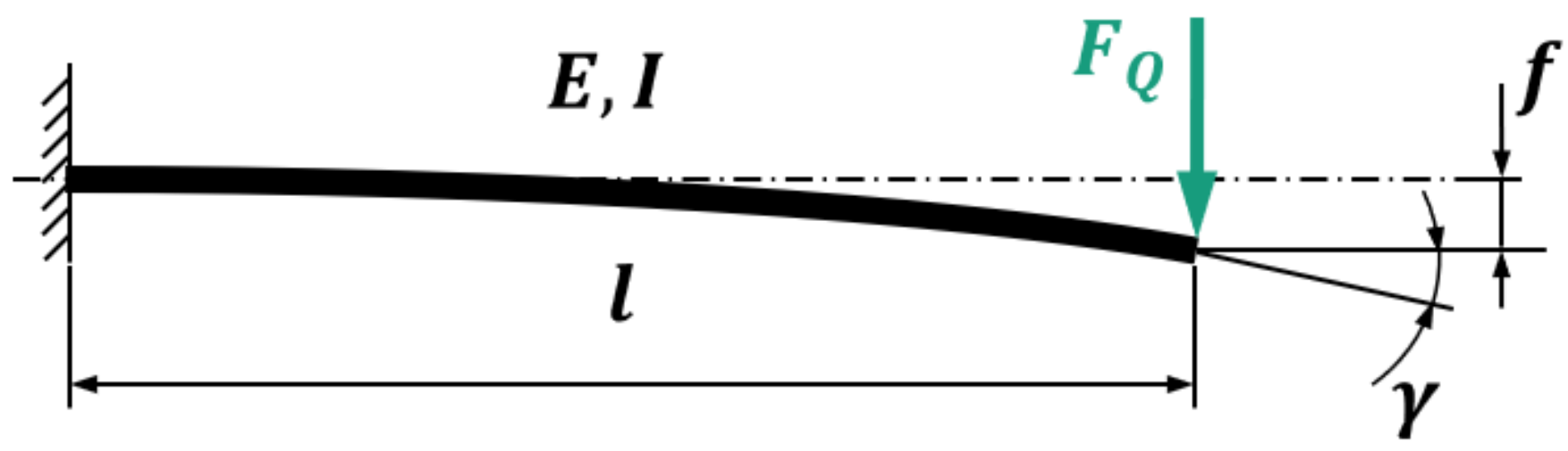}
    \caption{Unilaterally clamped bending beam and \snaphook (own illustration following \cite{classical.beam.models}).}
    \label{fig:4.2_Bending-Beam}
    \vspace*{-8pt}
\end{figure}

Developing the analytical joining models involves principles of classical mechanics, such as those provided in \cite{Gross.2017}. The \snaphook can be modelled as a unilaterally clamped, linearly elastic bending beam with length $l$, homogeneous mass distribution and modulus of elasticity $E$. For the relationship between the desired acting force $F=F_Q$ on the \snaphook of an electrical terminal (\figref{fig:4.2_Bending-Beam}) and the resulting deflection $f$ applies
\begin{equation}\label{eq:displacement-beam}
    f = \frac{F_Q \cdot l^3}{3 \cdot I_{y} \cdot E} \: ,
\end{equation}
where 
\begin{equation}\label{eq:slide-inertia-moment}
    I_{y} = \frac{b \cdot h^3}{12}
\end{equation}
is the area moment of inertia of the bending beam width $b$ and height $h$ against deflection. The inclination angle 
\begin{equation}\label{eq:inclination-angle-beam}
    \gamma \approx tan \; \gamma = \frac{F_Q \; \cdot l^2}{2 \; \cdot E \; \cdot I_y}
\end{equation}
of the bending line with respect to the horizontal axis is almost identical to the angular function $\text{tan} \: \gamma$ for small inclination angles.

\begin{figure}[!tb]
    \vspace*{8pt}
    \centering
    \includegraphics[width=3.5in]{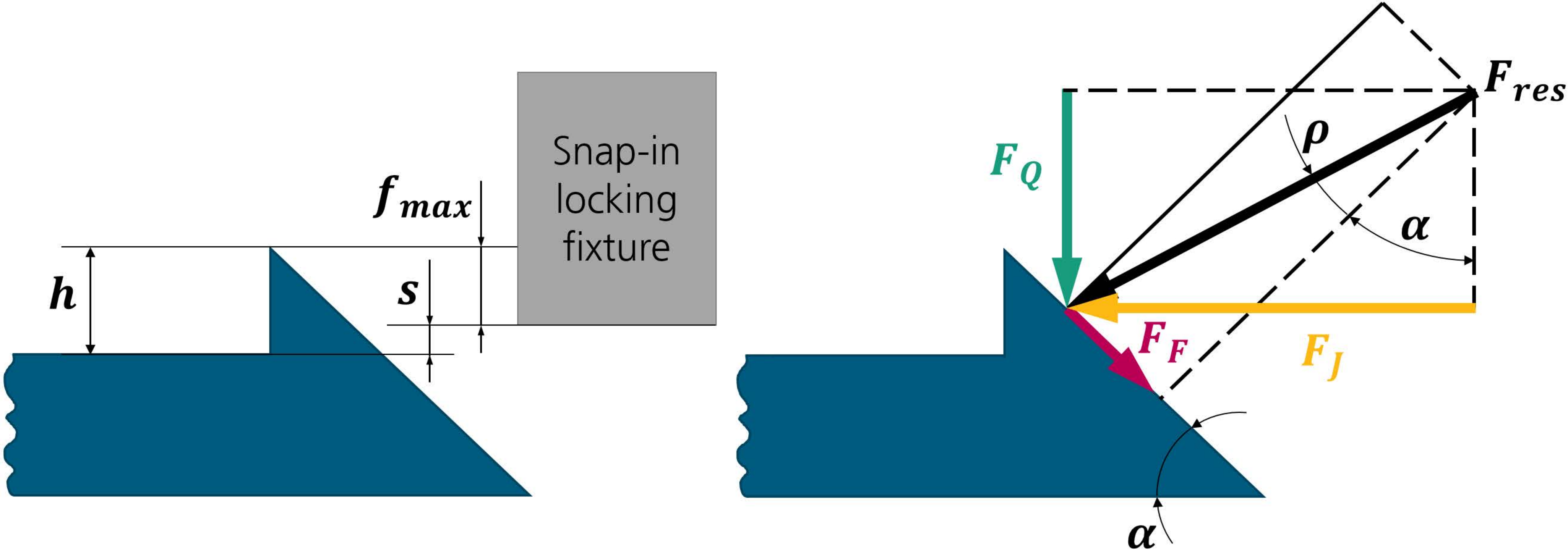}
    \caption{Geometric relations (left) and acting forces (right) on the \snaphook head.}
    \label{fig:4.2_Snaphook_displacement+acting-forces}
    \vspace*{-8pt}
\end{figure}

The maximum deflection of the bending beam $f_{max} = h_{k} - s$ occurs at the apex of the \snaphook $h_{k}$. The overlapping value $s$ can be directly calculated from the geometric relations between the \snaphook head and the snap-in locking fixture or the \DINrail (\figref{fig:4.2_Snaphook_displacement+acting-forces} left). For values $s < 0 \; \forall \; f_{max} > h_{k}$, the bending beam of the \snaphook is permanently deformed even after the joining process is completed.

In addition to the lateral force $F_Q$ deflecting the \snaphook, the joining force 
\begin{equation}\label{eq:Joining-Force}
    F_J = F_Q \; \cdot \text{tan}(\alpha + \rho)
\end{equation}
as well as the friction force $F_F$ act on the contact surface between the \snaphook head and the snap-in locking fixture (\figref{fig:4.2_Snaphook_displacement+acting-forces} right). The angle $\alpha$ is the effective joining angle, while 
$\rho$ results from the ratio of acting forces on the \snaphook head. Substituting the coefficient of friction $\mu_0 = \text{tan}(\rho)$ into \eqref{eq:Joining-Force} yields the relationship
\begin{equation}\label{eq:lateral-and-joinig-force}
    F_J = F_Q \; \cdot \frac{\mu_0 + \text{tan}(\alpha)}{1-\mu_0 \; \cdot \text{tan}(\alpha)}
\end{equation}
between the joining force $F_J$ and the lateral force $F_Q$ depending on the coefficient of friction.

Following the results of Kunz \cite{Kunz.2000}, the effective joining angle $\alpha$ is not a constant value but instead changes $\alpha' = \alpha + \gamma$ dynamically with the angle of inclination of the bending beam. Substituting in \eqref{eq:lateral-and-joinig-force} yields 
\begin{equation}\label{eq:forces-analytical-simpel}
    F_J = F_Q \; \cdot \frac{\mu_0 + tan(\alpha')}{1-\mu_0 \; \cdot tan(\alpha')} = F_Q \; \cdot \frac{\mu_0 + tan(\alpha + \gamma)}{1-\mu_0 \; \cdot tan(\alpha + \gamma)}
\end{equation}
for the acting joining force. The same applies to the effective length of the bending beam $l'$, which depends on the current point of force application and thus on the process progress and joining distance. Applying \eqref{eq:inclination-angle-beam} yields
\begin{equation}\label{eq:lateral-force-simple}
    \gamma' \approx \frac{F_Q \; \cdot l'^2}{2 \; \cdot E_S \; \cdot I_y} = \frac{3 \; \cdot f}{2 \; \cdot l'}
\end{equation}
for the actual inclination angle $\gamma'$, the displacement and the lateral force.
Finally, the material properties of the polymer of the \snaphook and the electrical terminal strongly depend on the present environmental conditions, such as moisture content and temperature. Therefore, the secant modulus $E_S$ is employed in the developed joining models instead of the commonly used modulus of elasticity $E$.

\subsection{Rigid Body Joining Models}\label{subsec:model-rigid}

Rigid body joining models~\cite{Laemmle.2022} are models that solely utilize available features of MuJoCo and build on classical mechanics. Like MuJoCo, most currently available off-the-shelf physics engines are rigid multi-body simulators at their core; they utilize sets of arbitrary geometries (rigid bodies) and constraints on their degrees of freedom (DoFs) with respect to one another or a reference such as a world coordinate frame. In contrast, the \snaphook in a terminal assembly is deformable; therefore, the elementary step to incorporate it into a MuJoCo simulation is to separate it into a set of rigid bodies and constraints. This approach aims to achieve a digital model of deformation that can accurately reproduce real-world scenarios by precisely modeling deformation dynamics. Specifically, this work decomposes the deformable part into a finite number of serially linked mass-spring-damper sub-models. One advantage of this structure is that the sub-models' parameters correspond well with those of classical beam models. Therefore, their values are usually available from the producer of the terminal assembly components as these beam models are used during the design process.

The parameters of a mass-spring-damper model are the mass, spring stiffness, and viscous damping. MuJoCo can calculate the mass and inertial properties of the parts based on their geometry and material density which are usually available from the respective CAD and supplier data, which leaves the stiffness and damping to be considered. Therefore, the following sections focus on presenting our calculations for these parameters. Models with one translational, one rotational, and two rotational DoFs are developed and presented. The significance of modeling not just the displacement of the snap-hook's head but also its deflection angle is examined in comparing the different models; decomposing into more parts should result in a more accurate model. In the following, three different versions of the rigid body joining models are described.

\begin{figure}[bt]
    \centering
    \includegraphics[width=3.4in]{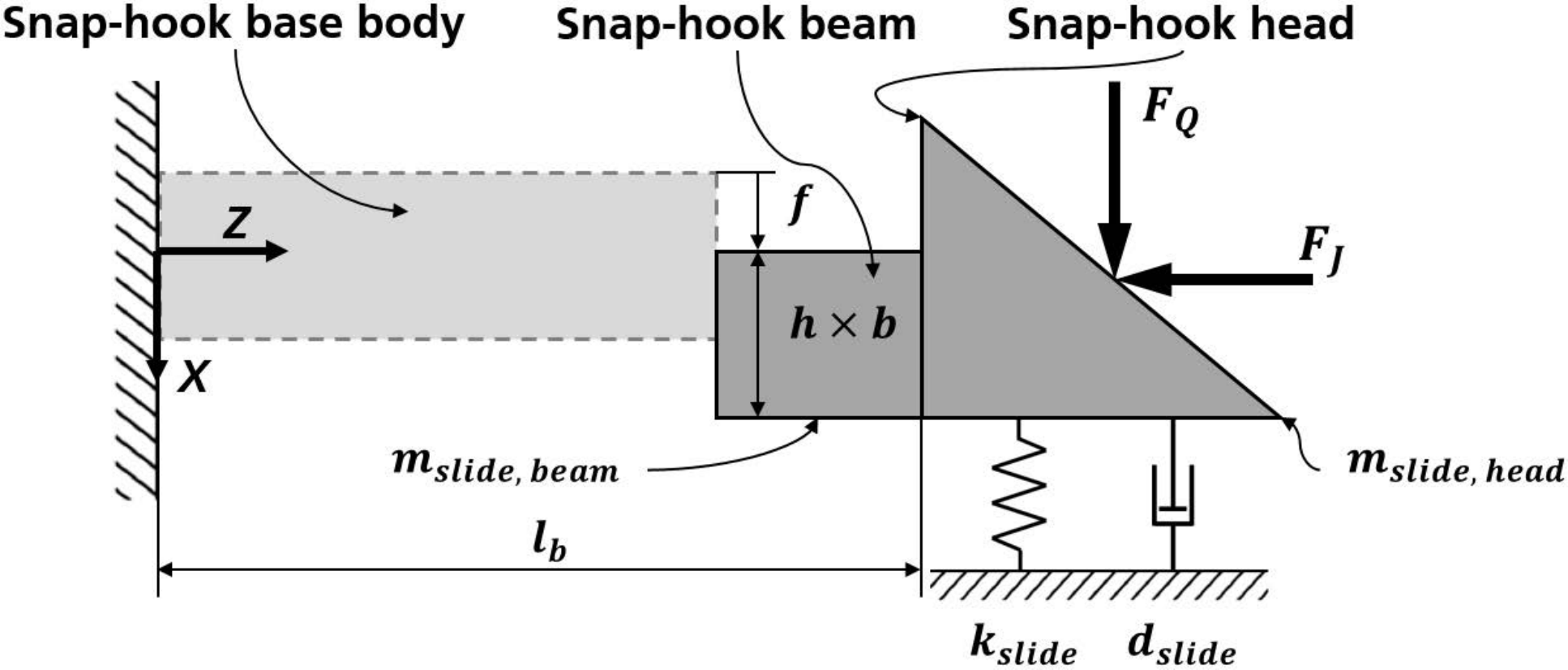}
    \caption{Rigid body joining model with one translational degree of freedom.}
    \label{fig:slide-model}
\end{figure}

\subsubsection{1-DoF Translational Model - "Slide Model"}

\figref{fig:slide-model} illustrates the model and its parameters. Should stiffness values be unavailable from the manufacturer of the electrical terminal part (through, e.g., the material's modulus of elasticity), an equivalent stiffness $ k $ against displacement $ f $ can be estimated according to \cite{classical.beam.models} as
\begin{equation}
    f \approx \frac{F_Q}{k} = \frac{F_Q \cdot {l}^3}{3 \cdot I \cdot E} \Longrightarrow k \approx \frac{3 \cdot I \cdot E}{{l}^3} ~,
    \label{eq:stiffness:slide}
\end{equation}
where $ F_Q $ is the acting lateral force on the deforming beam at distance $ l $ from its fixture along the beam's length, $ E $ is the modulus of elasticity, and $ I $ is the appropriate moment of inertia of the beam's cross-section.

We neglect the snap-hook head's dimension throughout the model parameterizations, thus assuming the point of load to be constant and at a full-beam length's distance from the snap-hook's base. These are apt assumptions in that the dimensions of the snap hook's head are usually small with respect to its length.

The viscous damping value is the joining model's degree of freedom. To our best knowledge, there is no baseline damping value for the presented use case; it should be tuned to achieve a stable simulation. To minimize the error of the dynamic system over its simulation time, its dynamics must be as fast as possible so the inaccurate phases during simulation are reduced. Dynamics with no overshoot furthermore avoid unintended resonance behavior, so assuming a coefficient of damping $ \xi\coloneqq1 $ and knowing each mass-spring-damper sub-model's mass $ m $ and stiffness $ k $, a reasonable estimate of the damping $ b $ can be determined and derived from a second-order mechanical system's equation of motion or transfer function according to
\begin{equation}
    b \approx b(\xi\!=\!1) = 2 \cdot \sqrt{\frac{m}{k}} \cdot k \cdot \xi \coloneqq \sqrt{4 \cdot m \cdot k} ~.\label{eq:damping}
\end{equation}
This approach aligns perfectly with the constraint dynamics' tuning in a MuJoCo model, which usually means setting a coefficient of damping and a dominant time constant after creating a model to define the error decay resulting from the constraint softness.

\subsubsection{1-DoF Rotational Model - "One-hinge Model"}

\begin{figure}[bt]
    \centering
    \includegraphics[width=3.4in]{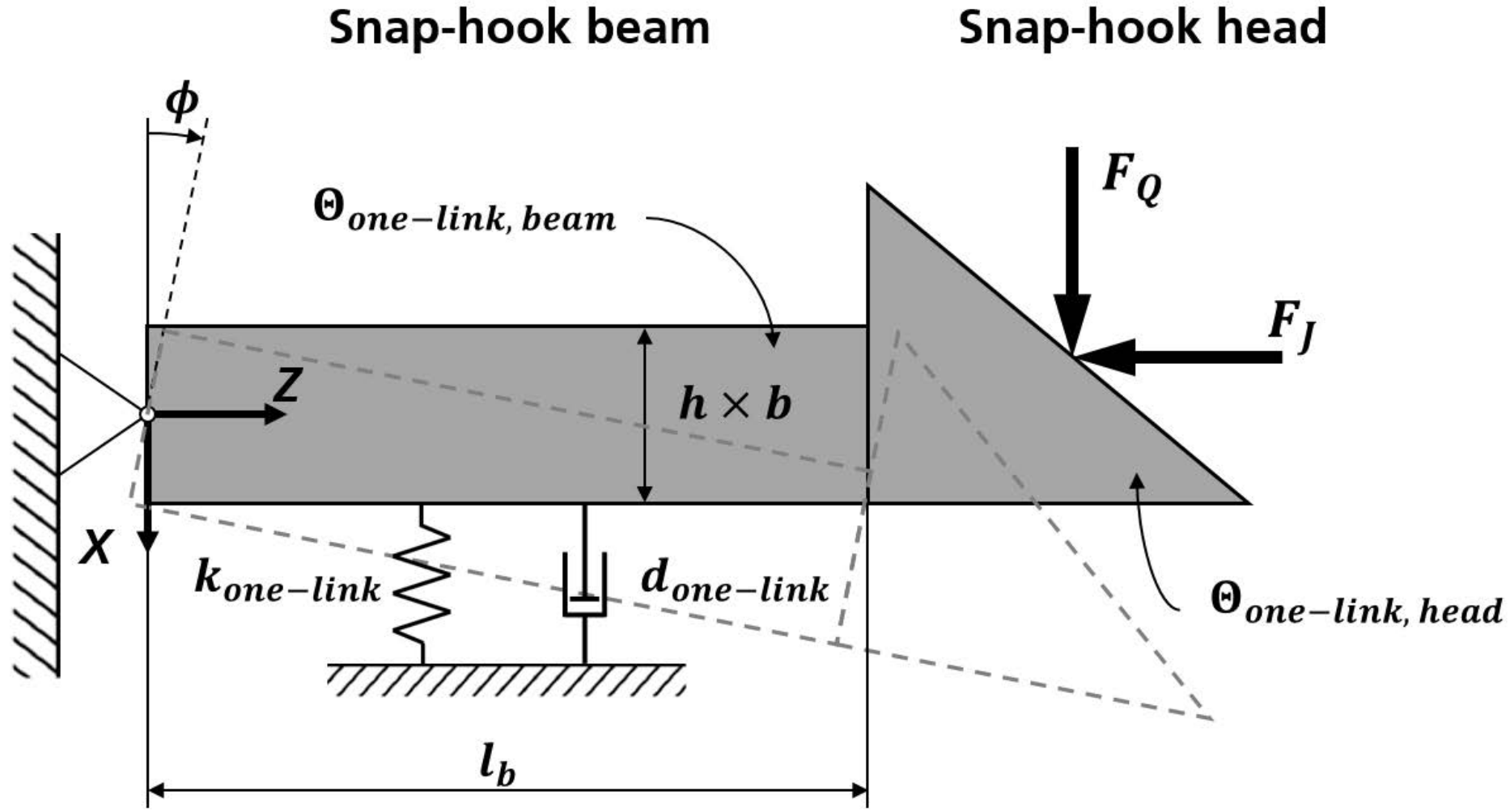}
    \caption{Rigid body joining model with one rotational degree of freedom.}
    \label{fig:onehinge-model}
\end{figure}

\figref{fig:onehinge-model} illustrates the model and its parameters. For the models with rotational DoFs, the equivalent stiffness against rotation $ k_t $ per its definition is
\begin{equation}
    k_t \coloneqq \frac{M_Q}{\varphi} \approx \frac{F_Q \cdot l}{\varphi(F_Q)} \approx \frac{\varepsilon \cdot l}{\atantwo \left\{ f(F_Q\!=\!\varepsilon),\, l \right\}} ~,
    \label{eq:stiffness:hinge}
\end{equation}
where $ M_Q $ is the torque load on the snap-hook that results from $ F_Q $ acting with a moment of arm $ l $, $ \varphi $ is the angular deformation due to the load, $ f(F_Q) $ is the translational displacement according to \eqref{eq:stiffness:slide}, and $ \varepsilon $ is a small perturbing force.

A MuJoCo model only allows static values for its parameters. However, the stiffness against rotation is a function of the force $ F_Q $, and to solve \eqref{eq:stiffness:slide}, $ F_Q $ must not equal $ 0 $. Since we expect small deformations and loads, we choose $ \varepsilon \coloneqq 10^{-7}\ N $ to avoid the singularity. Accordingly, our models are most accurate in the initial part of the assembly. The viscous damping can be calculated according to \eqref{eq:damping} using the equivalent stiffness $ k_t $.

\subsubsection{2-DoF Rotational Model - "Two-hinge Model"}

\begin{figure}[bt]
    \centering
    \includegraphics[width=3.4in]{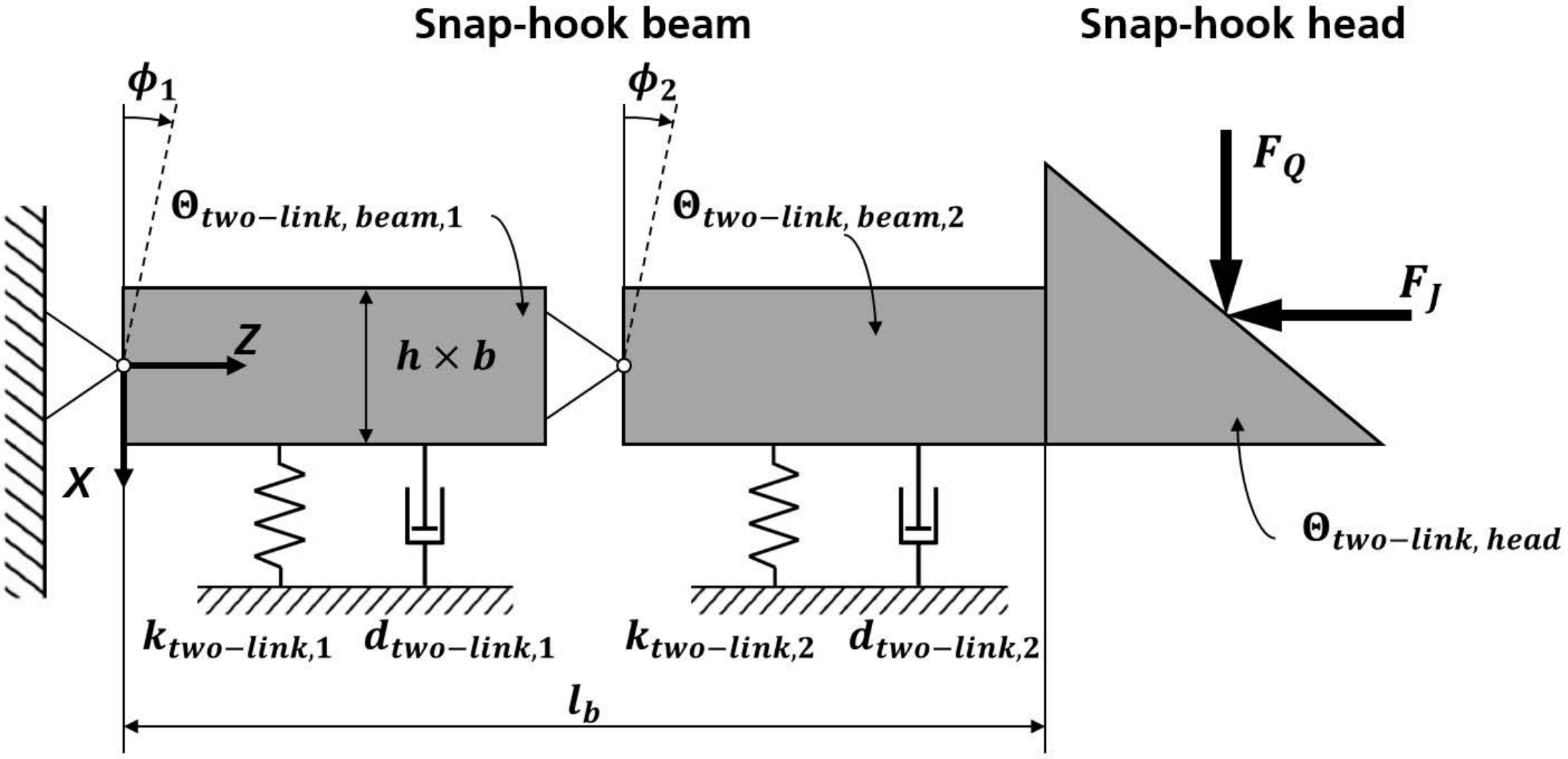}
    \caption{Rigid body joining model with two rotational degrees of freedom.}
    \label{fig:twohinge-model}
\end{figure}

\figref{fig:twohinge-model} illustrates the model and its parameters. For more than one serially linked mass-spring-damper, we calculat the resultant stiffness $ k_{res} $ of the system to be an equivalent one $ k_{equiv} $, e.g., following \eqref{eq:stiffness:slide}. The stiffness $ k_{res} $ of $ n $ similar springs, each with stiffness $ k $, is
\begin{equation}
    k_{res}^{-1} = n \cdot k^{-1} \rcoloneqq k_{equiv}^{-1} ~.
\end{equation}
Thus, the stiffness $ k $ of a mass-spring-damper sub-model in a series of $ n $ is
\begin{equation}
    k \coloneqq n \cdot k_{equiv} ~,
    \label{eq:spring-series}
\end{equation}
which holds for both stiffnesses against displacement and rotation. The viscous damping can be calculated according to \eqref{eq:damping} using the sub-model stiffness $ k $.

\subsection{Experimental Model Evaluation}\label{subsec:model-eval}

\begin{figure}[bt]
    \centering
    \includegraphics[width=3.4in]{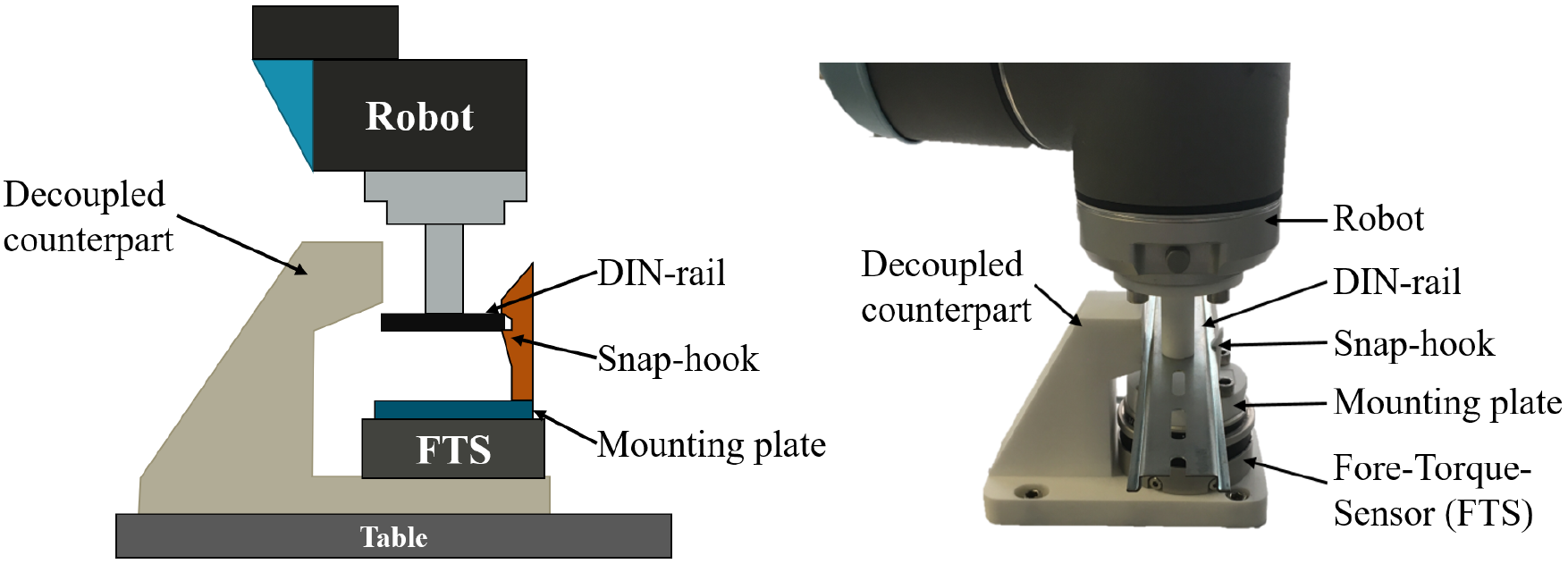}
    \caption{Experimental setup for measuring the acting forces on the \snaphook.}
    \label{fig:model-eval-setup}
\end{figure}

\begin{figure}[!t]
    \centering
    \includegraphics[width=3.4in]{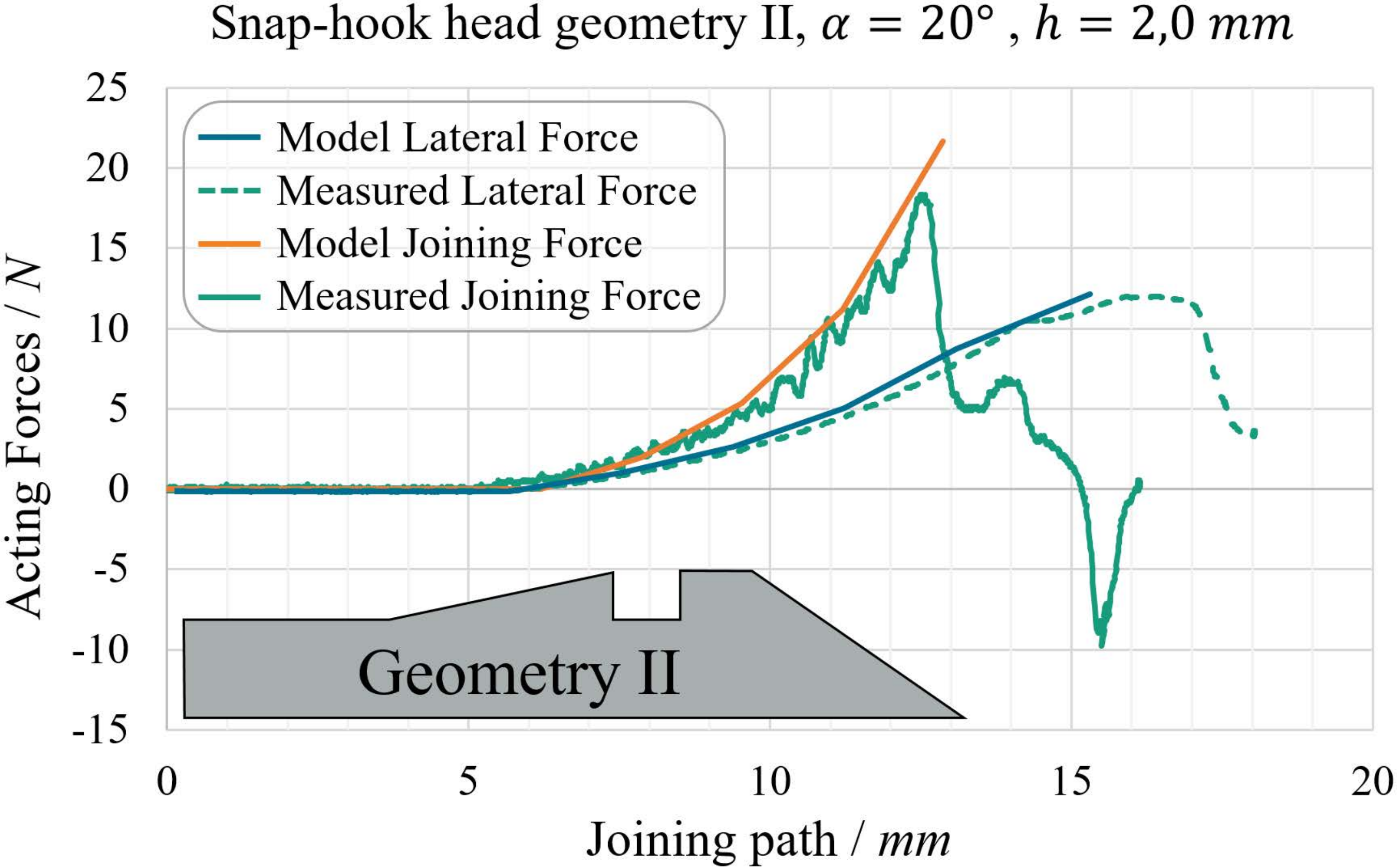}
    \caption{Results from the experimental model evaluation and force progression for a \snaphook with Geometry II.}
    \label{fig:analytical-model-results}
\end{figure}

For the comparison of the lateral and joining forces approximated with the joining models, a modified version of the physical robot test-bed described in Section \ref{sec:Real-Test-bed} is used. A specialised test setup for measuring the acting forces is additionally used to investigate a wide variety of different \snaphook geometries (\figref{fig:model-eval-setup}). The test setup consists of a conventional industrial force-torque sensor (FTS) that is permanently attached on one side to the table of the robot cell. An adapter plate is mounted on the opposite side of the FTS, which permits the straightforward insertion and exchange of different \snaphooks. By mechanically decoupling the two sides of the FTS, the forces acting on the \snaphook can be measured independently. During the experiment, the robot carries out the joining movement and deforms the \snaphook with the \DINrail attached to its flange. A fixed counterpart on the experimental setup provides the robot with additional guidance. In contrast to the usual automated assembly of electrical terminals, in the described setup, the \DINrail is moved by the robot and not the terminal block. This type of setup facilitates the accessibility of the \snaphooks and allows them to be easily exchanged for various experiments. 
The geometric properties of the \snaphooks used are varied for the evaluation of the analytical joining models. A total of 27 combinations of three different \snaphook geometries and of the three joining angles $\alpha = [20 \: ^\circ,30 \: ^\circ, 40 \: ^\circ]$ and three heights of the \snaphook head $h = [1.5 \: \textrm{mm},2.0 \: \textrm{mm},2.5 \: \textrm{mm}]$ are examined. The three evaluated \snaphook geometries differ significantly in the contour of the \snaphook. Geometry I has a directly declining edge after the maximum height of the \snaphook $h_{k}$, while geometries II and III initially cause a constant deflection with $h_{k}$ (see \figref{fig:analytical-model-results}). Compared to Geometry II, the deflection in Geometry III does not drop sharply but decreases linearly.
All \snaphooks used are produced from polyamide 12 using selective laser sintering. In the following, first, the developed analytical and, subsequently, the rigid-body joining models are evaluated.

Across all experiments, the calculated joining force and the measured joining force correspond closely. Furthermore, a qualitative consistency of the force progression is given for both the joining and lateral forces. At the same time, the approximated lateral forces exceed the measured lateral forces by a factor of 1.9 to 2.1. The deviations in calculating the lateral force can likewise be observed in evaluating the rigid-body joining models. The deviation factor in the approximated lateral forces is almost constant across all experiments, suggesting a systematic error. Possible explanations for the observed deviations lie in the model parameters, position deviations, and low stiffness of the lightweight robot used. In addition, the moisture content of the polymer used significantly influences the forces acting during the deflection of the \snaphook. Subsequent experiments should therefore investigate the forces acting on the \snaphooks under defined temperature and moisture conditions.

The quality of the lateral force calculation can be increased relatively straightforwardly by adding a pre-factor to Equation \eqref{eq:lateral-force-simple}. Incorporating the results from the experimental evaluation, the adapted calculation of the lateral force $F^*_Q$ thus yields

\begin{equation}\label{eq:lateral-force-simple-opt.}
    F^*_Q = \frac{1}{2} \; \cdot \frac{3 \; \cdot E_S \; \cdot I_Y  \; \cdot f}{{l'_{b}}^3}.
\end{equation}

\figref{fig:analytical-model-results} exemplarily presents the comparison of the approximated and measured lateral force after optimised calculation for a \snaphook with Geometry II, a joint angle $\alpha = 20 \: ^\circ$ and a head height of $h = 2.0 \: \textrm{mm}$. The qualitative and quantitative agreement of the calculated and measured forces is clearly visible. Furthermore, the mean absolute error of the forces is below $3 \: \textrm{N}$ for all investigated \snaphooks and thus, within the desired accuracy level for the subsequent training of the robot agent in the simulation.\\

For the investigation and evaluation of the rigid body joining models, we reproduced a real-world snap-fit assembly process in a virtual MuJoCo environment, incorporating the proposed rigid body joining models and integrating the analytical joining model using both the overlap-based approaches double ray measurement (DRM) and transformation matrix (TFM) from \cite{Laemmle.2022} using the official Python bindings of the physics engine MuJoCo.

\begin{figure}[!b]
    \centering
    \vskip 10pt
    \includegraphics[width=3.4in]{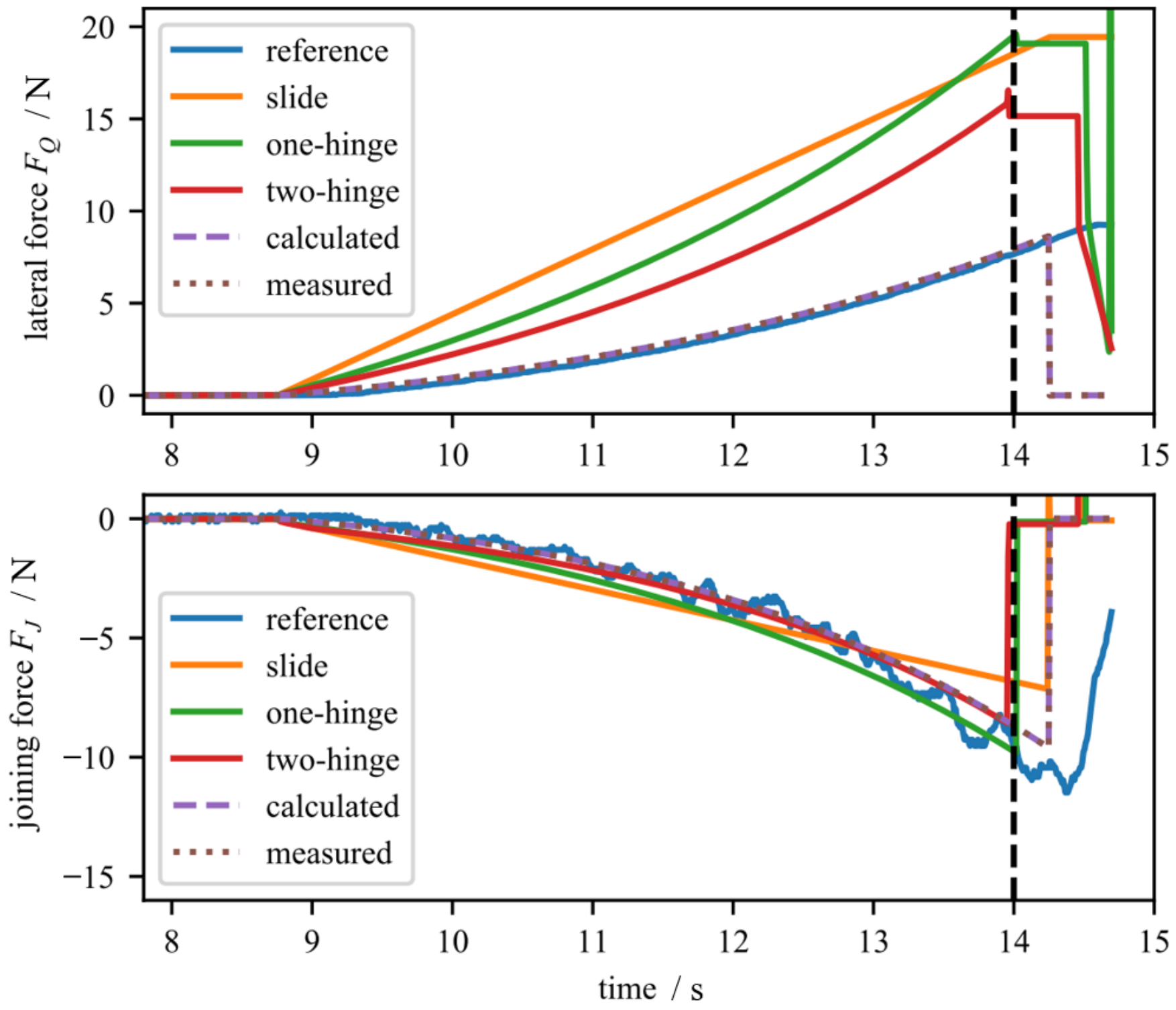}
    \caption{Lateral force (top) and joining force (bottom) values as recorded in the real-world scenario (\emph{reference}) and provided by the rigid body joining models (1-DoF translational model "\emph{slide}", 1-DoF rotational model "\emph{one-hinge}",  2-DoF rotational model "\emph{two-hinge}") and the overlap-based extensions DRM (\emph{measured}) and TFM (\emph{calculated}). The vertical dashed line marks the point of the snap-in, i.e., the point until the models are considered valid.}
    \label{fig:rigid-body-results}
\end{figure}

\figref{fig:rigid-body-results} illustrates the generated lateral and joining forces employing the rigid body joining models and the forces recorded from the real world robot test-bed. Apart from the acting forces, we also measured the models' \emph{real-to-simulation time} (real/sim) ratio. A value $ >\!1 $ means that the simulator can run the assembly with the given model faster than in real-time. Correspondingly, a value $ <\!1 $ means that the simulation of the process takes longer than executing its real-world counterpart would. 

Comparing the forces provided by the model with one translational DoF to that of one rotational, it is apparent that employing a model that also approximates the deflection angle results in more accurate synthetic data. The forces provided by the model with two rotational DoFs correspond better to the reference than those by the model with one. This tendency is as expected: a finer decomposition should yield results closer to the real-world phenomenon. However, the hardware bottleneck is significant: the real/sim ratio of the model with two rotational DoFs is $ 0.5768 $, while that of the models with only one DoF are $ 1.229 $ and $ 1.108 $ for the rotational and translational models, respectively.

The TFM and DRM approaches produce practically the same accurate joining forces throughout the simulation as the rigid body joining model with two rotational degrees of freedom. They, however, allow the decoupling of the joining and lateral forces from one another, resulting in more accurate lateral forces. The calculation and measurement approaches produce the same forces because both can reproduce the snap-hook head's actual profile and digitally represent the overlap definition well. However, the measurement method can be generalized better, as the calculation explicitly requires the snap-hook head's detailed geometry analytically.

These overlap-based methods show excellent potential also in mending the computational load of the rigid body joining models: the TFM and DRM approaches can run with real/sim ratios of $ 8.793 $ and $ 10.81 $, respectively. Neither has to represent deformation dynamics accurately but only map the simulation state to acting forces directly. Furthermore, when there is no overlap, the necessary computations can be short-circuited, further alleviating the computational load and resulting in real/sim ratios of $ 10.06 $ and $ 17.38 $ for the TFM and DRM methods, respectively.

As a caveat, both of the overlap methods require setting at least one of the parts in the assembly to be penetrable in the simulation. This modification can and usually does lead to behaviours that are not plausible in reality but are made possible in simulation. Without addressing this issue, an applied deep reinforcement learning method would generally exploit these loopholes and learn invalid robot control policies. We have considered the problem and developed countermeasures to avoid such behaviours with success by deriving the overlap over time. Discontinuities in the course of the overlap indicate the components' forbidden behaviour in the simulation, which, in turn, signals the agent exploiting the permitted intersection of geometries. Punishment through a negative reward apparently prevents unwanted behaviour and leads to more reliable learning. Nonetheless, sacrificing the inherent robustness of the simulator remains a caveat of custom extensions. Their application requires additional considerations and development, whereas the rigid body joining models are significantly more resistant to invalid exploitation by DRL methods.

\section{Simulation-based Learning}
\label{sec:learning}

The policies for selecting and parameterizing the \robotskills are trained in the simulation extended by the joining models introduced in the previous section. The learning environments are implemented according to Section \ref{subsec:RL-Framework}. For training the robot agent, the Stable Baseline 3 \cite{StabelBaselines3} implementation of the off-policy learning algorithm Soft-Actor Critic (SAC) is adopted. Subsequently, the modeled robot test-bed in the simulation is described first. Then the training results and their evaluation in the simulation are presented and finally discussed.

\subsection{Setup of the Simulated Robotic Test-Bed}\label{subsec:setup-sim}

\begin{figure}[!tb]
    \vspace*{8pt}
    \centering
    \includegraphics[width=3.4in]{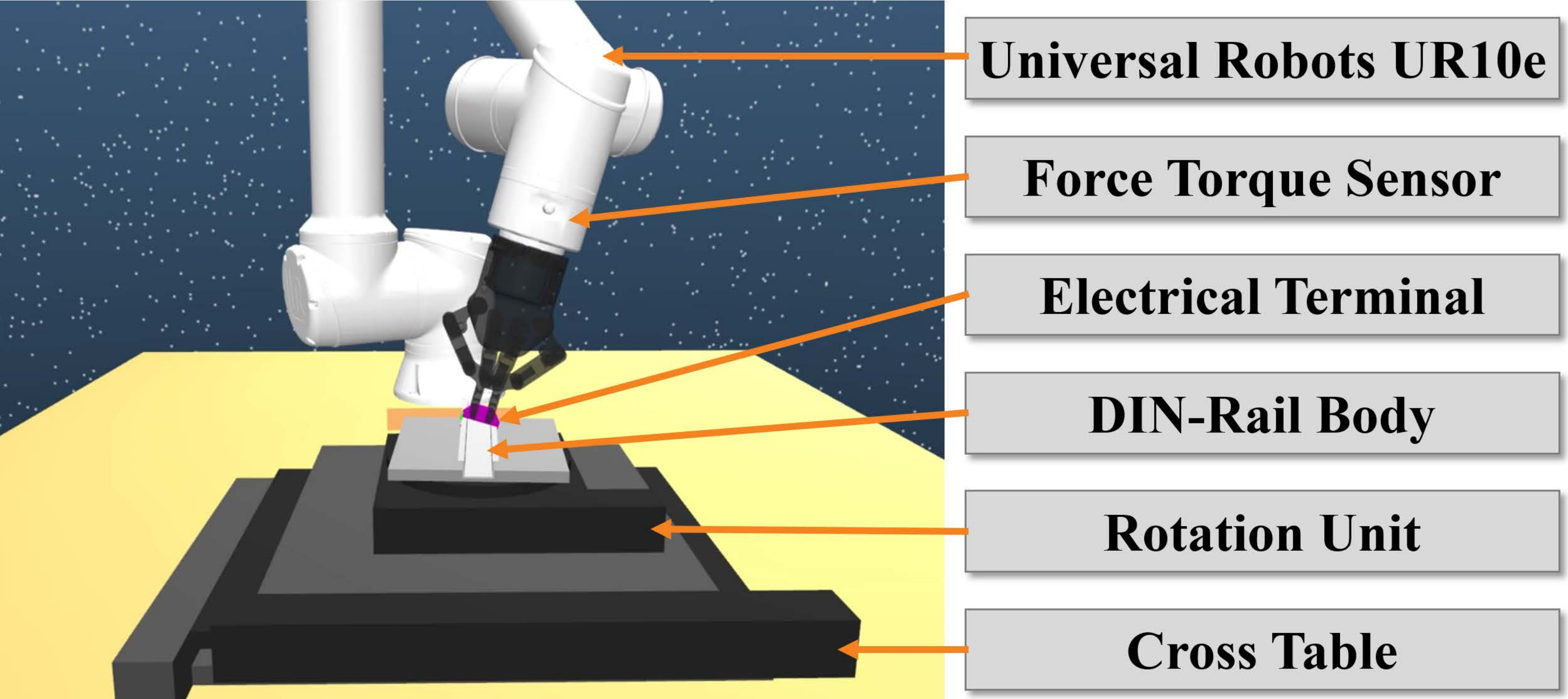}
    \caption{Simulated model of the robotic test-bed including the integrated joining models.}
    \label{fig:5.1_Simulated_Test-bed}
    \vspace*{-8pt}
\end{figure}

Both the simulated and the physical test-bed employ a Universal Robots UR10 e-series lightweight manipulator with an integrated force-torque sensor (FTS) at its flange and a gripper mounted (\figref{fig:5.1_Simulated_Test-bed}). In the simulation, the part held is attached rigidly to the gripper fingers along its rotational degrees of freedom and through soft equality constraints along the Cartesian translational ones. Similarly to previous work, these constraints are necessary to enable the constraint solver of the used physics engine. They are tuned through their second-order error decay dynamics so that their effect is negligible on the controller's observations, i.e., their error decays 10 times faster than the controller observes the environment. By doing so, our controller can access the simulated manipulator's joint velocities directly without resulting in "jumpy" robot movements and unrealistically high forces, as the velocities would ordinarily be outputs of a simulation step.

\subsection{Training Results in Simulation}\label{subsec:results-sim}

At the beginning of each training episode, the position and orientation of the modeled \DINrail are randomized. A training episode terminates once the robot agent has successfully assembled the terminal on the \DINrail, the available $N=6$ skills per episode have been exhausted, or the agent has exceeded the maximum permissible forces. To avoid systematic influences of the domain randomization, a total of five training runs with different \textit{random seeds} are executed simultaneously. In each training run, the robot agent has up to $n=\num{100,000}$ skill executions at its disposal, with a time limit of $10.0 \: \textrm{s}$ for each skill. During training, the policies are evaluated after performing $1,000$ \robotskills. Thus, each evaluation includes $100$ policy executions.

\begin{figure}[!tb]
    \vspace*{8pt}
    \centering
    \includegraphics[width=3.5in]{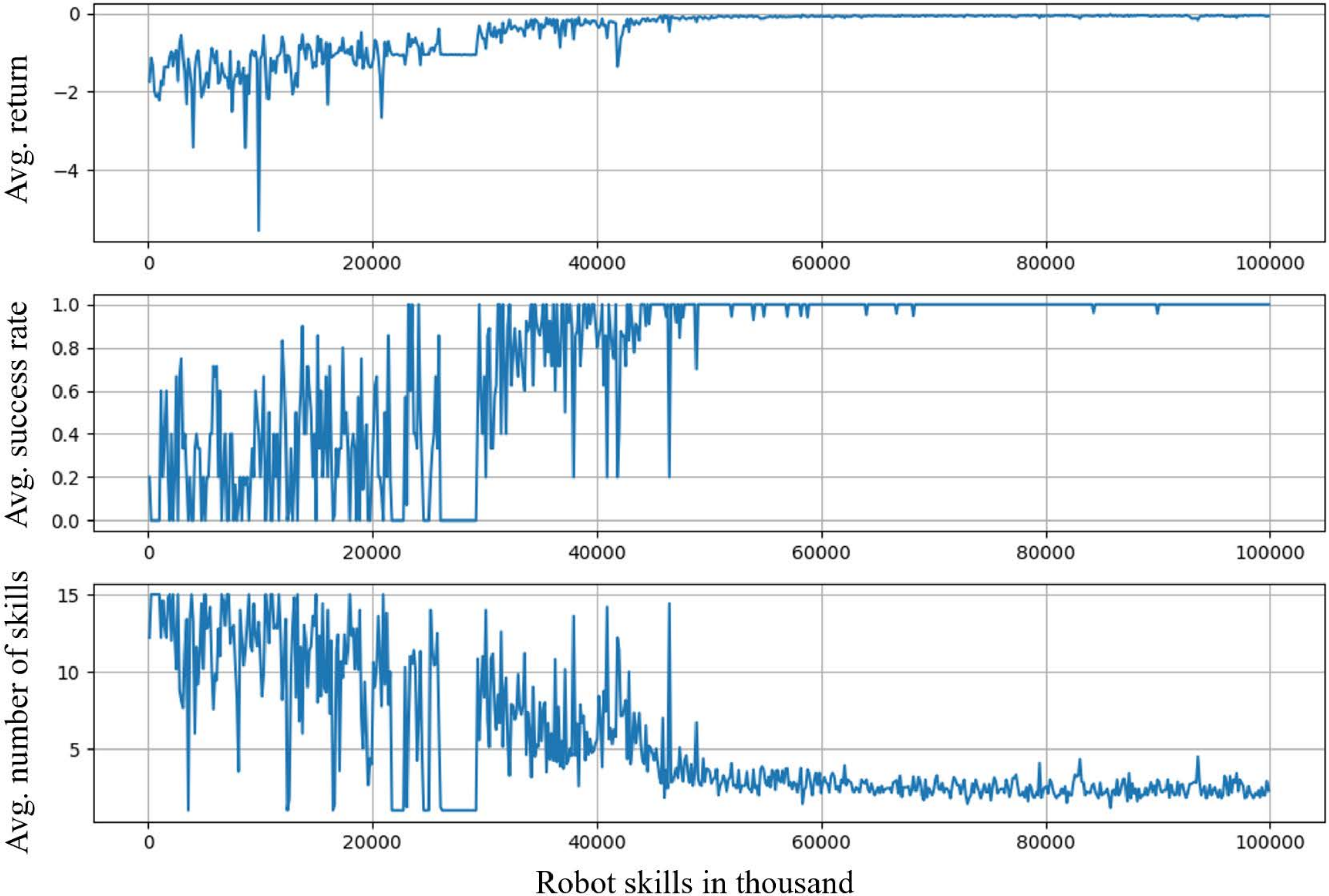}
    \caption{Average return, success rate and number of used skills from the simulation-based training of a policy trained with SAC.}
    \label{fig:5.2_Simulation_13c_SAC_20220511_112141}
    \vspace*{-8pt}
\end{figure}

\figref{fig:5.2_Simulation_13c_SAC_20220511_112141} illustrates the typical development of the average return, success rate, and the number of skills used per episode of a policy trained with SAC during the evaluation parallel to the training. The average return converges towards the optimum of $0$ and the average success rate towards the maximum of $100 \%$. Similarly, a convergence of the average skills required towards the minimum of two skills can be seen. Especially in the first half of the evaluation, strong oscillations in the evaluation results are still visible. However, starting at about $45,000$ skill executions onwards, these oscillations decrease significantly, which can be attributed to the increasing learning success of the agent (cf. Section \ref{subsec:discuss-sim}).

In addition, five evaluation experiments were conducted with fully trained policies. To test the robustness of the in-silico trained policies, the orientation of the \DINrail was additionally randomized significantly beyond industry-typical inaccuracies. The orientation is changed in $2 \textrm{°}$ steps within the interval $[-8 \textrm{°},8\textrm{°}]$. Furthermore, four supplementary positions of the terminals were evaluated at a distance of $[-60 \: \textrm{mm},-30 \: \textrm{mm}, 30 \: \textrm{mm}, 60 \: \textrm{mm}]$ from the ideal trained position at $0 \: \textrm{mm}$ along the longitudinal axis of the \DINrail. These additional four positions were not considered during training and are thus new to the robot agent. Finally, the generalization ability of the trained behavior is evaluated. For this purpose, a policy is executed with a different terminal than during training. 

\begin{figure*}[!tb]
    \centering
    \includegraphics[width=\textwidth]{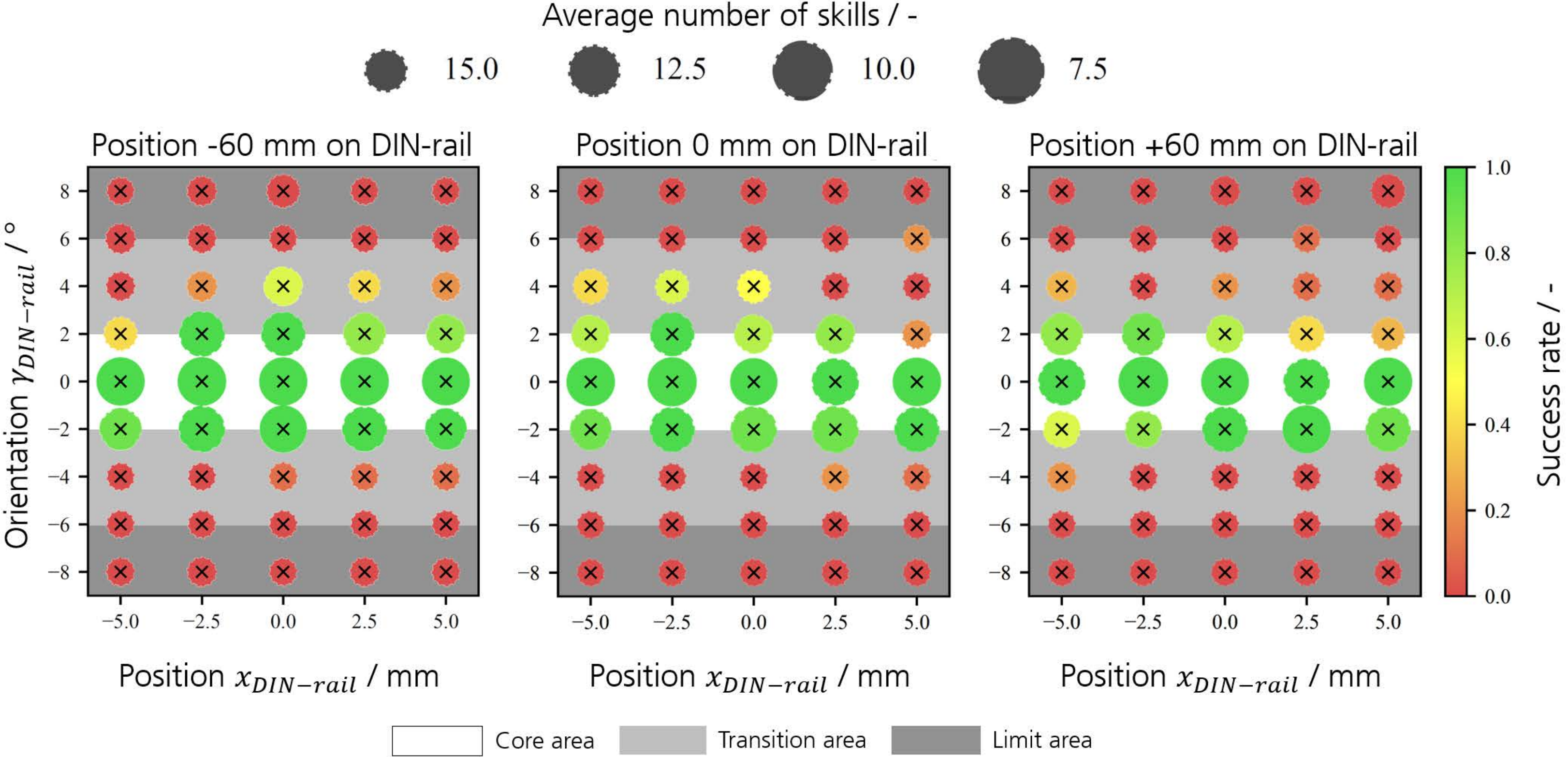}
    \caption{Success rates for a policy trained and evaluated on the WAGO 2002-1201 terminal during in-silico execution.},
    \label{fig:5.2_Simulation_Evaluation-WAGO-2002-1201}
    \vspace*{-8pt}
\end{figure*}

The evaluation results of a policy trained for and evaluated on the WAGO 2002-1201 terminal are presented in \figref{fig:5.2_Simulation_Evaluation-WAGO-2002-1201}. The agent achieves success rates of up to $100 \%$ in the core area even for position deviations of the \DINrail of $\pm 5 \: \textrm{mm}$. In the transition area with orientations of the \DINrail of $\pm 2 \: \textrm{°}$, the agent still achieves success rates between $90 \: \%$ and $95 \: \%$. Thereby, deviations of the orientations in positive direction $\gamma_{\DINrail} < 0 \: \textrm{°}$ are compensated more robustly compared to the opposite direction. As expected, the success rate for larger rotations of the \DINrail decreases significantly up to the limit areas. At the same time, these ranges are industrially highly atypical and can be avoided by an appropriate design of the robot cell. Similar evaluation results are observed for the other four experiments, even for the transfer to a terminal not employed during training.

\subsection{Discussion of Simulation Evaluation Experiments}\label{subsec:discuss-sim}

The evident oscillations of the return, the success rate, and the required skills in the first half of the training can be attributed to the structure of the \elta task (\figref{fig:2_Eletrical-Terminal-Assembly_Process_Overview}). First, the agent needs to use the $\mathfrak{S}_{approach}$ and $\mathfrak{S}_{slide}$ skills to place the fixed hook of the terminal below the edge of the \DINrail before it can successfully execute the entire process. Once the agent has confidently learned this first half of the assembly process, it can subsequently mount the deformable \snaphook by executing the pivot skill $\mathfrak{S}_{pivot}$. Starting at this level at approximately $45,000$ \robotskills executed in \figref{fig:5.2_Simulation_13c_SAC_20220511_112141}, the agent consistently achieves success rates approaching the $100 \: \%$ limit. In addition, the agent can minimize the number of skills required in the further course of the training.

Across all evaluation experiments, the agent displays a high degree of robustness to translational displacements of the \DINrail. Furthermore, the agent successfully applies its extended action space in a goal-directed manner. Suppose the agent does not reach its goal with the first skill execution $\mathfrak{S}_{terminal,1}$. In that case, it adjusts the initial position of the terminal in the first skill $\mathfrak{S}_{lin}$ of the second skill execution $\mathfrak{S}_{terminal,2}$ to execute the process successfully subsequently.
Similarly, it is also possible for the agent to adjust rotations of the \DINrail in the core areas by single pivot skills $\mathfrak{S}_{pivot}$, even if the terminal has already been snapped. The expected lower success rates for rotations of the \DINrail outside the core areas can be attributed to a natural limit of force-controlled skills and the success criteria used. The agent can only draw limited conclusions about the actual rotation of the \DINrail from the measured forces. At the same time, the defined success condition only classifies the assembly as a success if the assembled terminal is less than $\pm 1 \: \textrm{°}$ distorted on the \DINrail. Wide terminals with multiple bending beams are advantageous in this respect due to the higher stiffness compared to narrow terminals with a single bending beam. The combination of force-controlled skills with camera systems and image processing offers an appropriate option if rotations outside of the core areas are also to be mounted reliably.
The preferred direction of the agent towards negative angles of rotation of the \DINrail can be attributed to a recess on one side of the \snaphook head of the terminal. This facilitates the agent's ability to mount the terminal at rotation angles with $\gamma_{\DINrail} < 0 \: \textrm{°}$. Favorably, this effect is amplified at positions $[-60 \: \textrm{mm}, -30 \: \textrm{mm}]$ along the longitudinal axis of the \DINrail.

Finally, the robot agent also successfully generalizes to terminals that were not considered during training, with success rates up to $100 \: \%$. As expected, the transfer to new terminals is limited if the terminal to be executed requires significantly higher assembly forces than were learned during the training.
\section{\SimtoReal Transfer}
\label{sec:evaluation}
To evaluate the agent's trained behaviour, the trained policies are transferred to a physical robot system. First, the structure of the robot system used is briefly presented. Then the evaluation experiments carried out are described and finally discussed. It should be noted that the policies are performed in reality without any further training.

\subsection{Setup of Real-World Robotic Test-Bed}\label{subsec:setup-real}
\label{sec:Real-Test-bed}
The physical robot system employs a Universal Robots UR10e series lightweight robot. A CRG gripper from Weiss Robotics is attached to the robot's integrated force-torque sensor. The mounted gripper clamps were designed for terminals from WAGO and manufactured using an SLA printing process. The terminals are gripped automatically with manually implemented skill routines, as they are also employed in industrial environments. We deliberately refrain from using force-controlled skills to evaluate realistic inaccuracies and tolerances through the gripping process. The communication between the skill framework, the robot controller and all peripheral devices is handled by the Robot Operation System (ROS). At the beginning of a policy roll-out, the policy selects the first skill and sets its parameters. Subsequently, the skill is executed by the robot. If the skill has been completed, but the process goal has yet to be reached, the policy selects and parameterises an additional succeeding skill. This alternating procedure continues until the assembly is successfully completed, the maximum number $N=6$ of skills has been reached, or one of the stop conditions has been triggered. 
The position and orientation of the \DINrail are set through a mechanical positioning unit consisting of a cross table and a rotation unit. Before starting an evaluation, the positioning unit is calibrated to minimise potential distortions of the evaluation results. The robot and the positioning unit are mounted on a welding table in the cell. Due to the precision of the positioning elements on the welding table, a high reproducibility of the mechanical construction of the cell is achieved. Finally, the robot cell is completed by a perspex enclosure to protect the human operator from malfunctions during execution. Thus, the simulated robot test bed in the source domain is a realistic replica of the physical robot cell in the target domain. Only non-process-relevant components, such as the safety devices used, were not modelled in the simulation.

\subsection{Execution Results in Reality}\label{subsec:results-real} 
The three experiments in \tabref{tab:elta-real-evaluation-policy-overview} were defined for the evaluation. Analogous to the evaluation in the simulation, in each evaluation experiment, the translational position of the \DINrail is adjusted in $2.5 \: \textrm{mm}$ steps within the interval $[-5.0 \: \textrm{mm},5.0 \: \textrm{mm}]$. The orientation is changed in $2.0 \: \textrm{°}$ steps within the interval $[-8.0 \: \textrm{°},8.0 \: \textrm{°}]$. In experiment I and experiment II, five terminal positions are also varied in $30.0 \: \textrm{mm}$ steps along the \DINrail. Due to the wider terminal in experiment III, four positions were evaluated at $[-60.0 \: \textrm{mm}, -20.0 \: \textrm{mm}, 20.0 \: \textrm{mm}, 60.0 \: \textrm{mm}]$. Each experimental run is repeated once. Thus, a total of 1034 assembly processes were carried out in the three experiments.

\begin{table}[!h]
    \centering
    \begin{tabular}{l|c|c}
        \hline
        \textbf{Policy}  & \textbf{Trained for} & \textbf{Evaluated on} \\
        \hline
        \textbf{Experiment I} & WAGO 2002-1201 & WAGO 2002-1201\\
        \textbf{Experiment II} & WAGO 2102-1301 & WAGO 2102-1301\\
        \textbf{Experiment III} & WAGO 2102-1301 & WAGO 2016-1301\\
        \hline
    \end{tabular}
    \vspace*{8pt}
    \caption{Overview of trained and evaluated policies on the physical test-bed.}
    \label{tab:elta-real-evaluation-policy-overview}
\end{table}

\begin{figure*}[!tb]
    \centering
    \includegraphics[width=\textwidth]{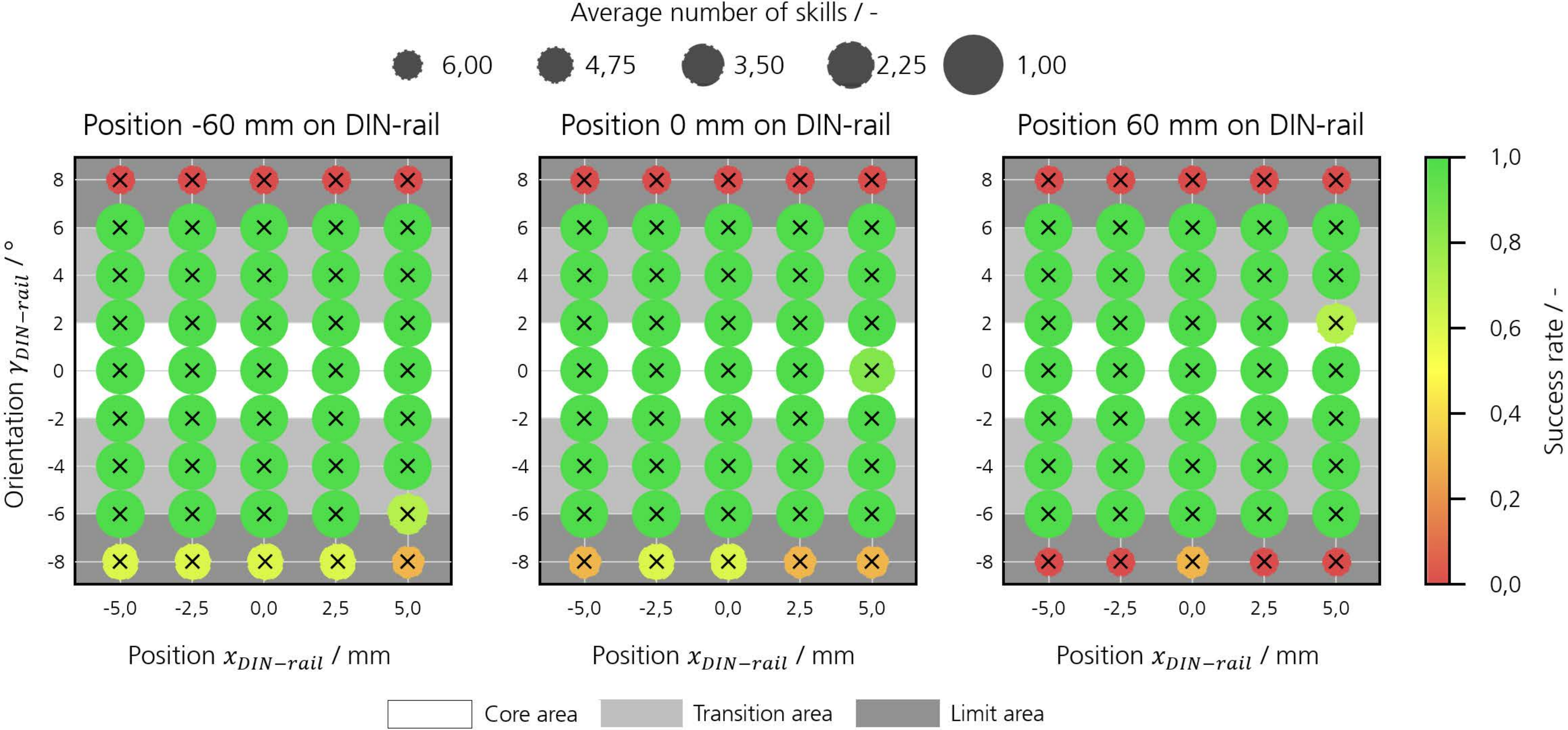}
    \caption{Success rates for a policy trained and evaluated on the WAGO 2002-1201 terminal during real-world execution.},
    \label{fig:6.2_Real_Training+Eval-WAGO-2002-1201}
    \vspace*{-8pt}
\end{figure*}

The results of the real-world execution of a policy trained and thus also evaluated for the WAGO 2002-1201 terminal in experiment I are shown in \figref{fig:6.2_Real_Training+Eval-WAGO-2002-1201}. The plot shows consistently high success rates of up to $100 \: \%$ in both the core and transition areas. The success rate decreases only for maximum rotations of the \DINrail of $\pm 8 \: \textrm{°}$. As in the simulation, the agent demonstrates robust behaviour against translational deviations of the terminal. With few exceptions, the robot agent succeeds in \elta with the minimum number of skills. Almost identical evaluation results are obtained for experiments II and III. Here, too, the agent achieves almost exclusively a success rate of up to $100 \: \%$ in the core and transition areas. Even in the limit areas, the agent succeeds in process-safe assembly for orientations of the \DINrail of $\pm 6 \: \textrm{°}$. The limit of the trained behaviour is for orientations of $\pm 8 \: \textrm{°}$. However, such large deviations are well above typical industrial inaccuracies.

\subsection{Discussion of Real-World Experiments}\label{subsec:discuss-real}

Across all evaluation experiments performed, the agent consistently achieves impressive success rates at the $100 \: \%$ limit in both the core and transition areas. With a limit of successful assembly processes at orientations $\pm 6 \: \textrm{°}$ of the \DINrail in the limit ranges, the agent thus clearly exceeds the expectations of industrially typical inaccuracies. All investigated translational inaccuracies of the \DINrail are successfully compensated by the agent through the trained skills or the use of additional skills. Observed outliers in the evaluation experiments were almost exclusively caused by a displacement of the terminal in the gripper. In the process, the terminal is pulled slightly out of the friction-locked grip. As a result, the relative position between the terminal and the gripper changes, and the agent needs an additional skill to compensate for the resulting deviation. The agent also compensates for minor deviations occurring during the regular gripping process. Only in rare extreme cases does the position and orientation of the terminal in the gripper change so significantly that assembly is no longer possible.

In the physical evaluation experiments, too, a preferred direction of the agent towards negative rotation angles $\gamma_{\DINrail} < 0 \: \textrm{°}$ of the \DINrail is recognisable (cf. \figref{fig:6.2_Real_Training+Eval-WAGO-2002-1201}). Once again, this effect can be attributed to the recess at the head of the \snaphook on the terminal.
An advantage for the high robustness against inaccuracies and the ability to generalise to new positions and terminals is undoubtedly the flexibility of the polymer of the terminals when executed on a physical robot system. As a result, the robot can also successfully compensate for higher deviations in the orientation of the \DINrail compared to the simulation. At the same time, the compliance of the terminals limits any compensating rotation of the mounted terminal on the \DINrail. The agent aims to align the terminal perpendicular to the longitudinal axis of the \DINrail. However, as already shown in the simulation, the agent can only derive limited conclusions about the current orientation of the terminal from the measured mounting forces. Stiffer terminals, such as the wide WAGO 2102-1301 with three bending beams, have an advantage over more flexible terminals, such as the narrow WAGO 2002-1201.

Finally, the observations from the simulation on the transfer of the learned behaviour to previously untrained terminals are also confirmed in reality. As expected, successful generalisation is possible if the required assembly forces during execution are not significantly lower than the trained forces. This allows, in particular, the transfer of policies trained for stiff terminals to more compliant terminals, while the opposite transfer is naturally limited.

\section{Conclusion}
\label{sec:conclusion}

In the subsequent chapter, we summarize and critically assess the research and development efforts undertaken. Additionally, we provide an outlook on related and future research topics.

\subsection{Summary of Contributions}\label{subsec:conclusion-sum}
The presented work involves training parametrisable \robotskills in a rigid body simulation for the force-controlled assembly of electrical terminals on a \DINrail. The research and development work focuses on two main areas: enhancing the physics simulation and modelling the snap hook assembly, and integrating the improved simulation with deep reinforcement learning algorithms to train force-controlled \robotskills. 

Analytical as well as rigid body joining models for deformable \snaphooks were developed, to determine the acting forces during assembly. For the evaluation of both types of joining models, a robot-based testbed was designed to measure the assembly forces. Finally, the joining models were integrated into the simulation.

To train policies as control algorithms for the robot, the two learning algorithms Soft-Actor Critic (SAC) and Twin Delayed Deep Deterministic Policy Gradient (TD3) were utilized.
Extensive validation experiments were conducted in the simulation as well as on the physical robot system, using various different electrical terminals. To investigate the generalization capability of the trained policies against typical process inaccuracies and tolerances, a mechanical positioning device was employed. Furthermore, policies were tested with terminals that were not used for their training.

\subsection{Critical Assessment and System Limitations}\label{subsec:conclusion-critical}

Both, the analytical as well as the rigid body joining models allow precise force determination on the \snaphook. After optimization, the analytical joining models present a mean absolute error of the forces below $3 \: \textrm{N}$ for both the lateral and the joining force. The rigid body joining models even outperform the analytical models for force determination on the joining force, while the model overestimates the lateral forces by a factor of 1.5 to 2.0. However, employing the rigid body joining models requires a very small time step in the simulation, reducing the possibility of training faster than real-time.


During evaluation in the simulation, the robot agent achieves success rates of up to $100\%$ in the core area, even with larger translational displacements of the \DINrail. Furthermore, the agent successfully uses the additional \robotskills available to compensate for the terminal block's rotational deviations. Only in the transition and limit areas do the success rates drop drastically. This is particularly due to how the success criteria is defined in the simulation.

After the transfer of the trained policies to the physical robot system, the agent outperforms the in-silico evaluation with success rates up to the $100 \: \%$ limit, even above tolerances typical for the industry. Thereby, the agent executes highly robust behaviour against translational displacements of the \DINrail in both the core and the transition area. The agent only showed a limit in the successful execution for strong rotations of the \DINrail above $8.0 \: \textrm{°}$. Furthermore, transferring trained policies for wide terminals on processes with narrow terminals works well, while the reverse represents a boundary.

\subsection{Outlook to Future Work}\label{subsec:conclusion-outlook}
Ongoing and future work focuses on promising advancements, particularly integrating hybrid learning methods. These methods will optimize both the parameters of the skills and their selection sequence. This progress will enable the training of complete robot programs without manual intervention and even create new skills through compositions.

Besides the learning environment and training, the developed joining models and the simulation environment present significant future research and development opportunities. Simplifying the parameterisation of the environment, for instance, using machine learning is a particularly promising approach. 

Finally, the joining models themselves can also be further examined and optimized, for example, for applications in the design process itself. In this case, the focus is set on the exact determination of the acting forces and the deformation behaviour of the \snaphooks. The development of a dedicated test bed for precise measurement of the acting forces is the subject of ongoing research work.
\section*{CRediT authorship contribution statement}

\textbf{Arik Lämmle}: Conceptualization, Methodology, Validation, Investigation, Writing - Original Draft, Writing - Review \& Editing, Visualization, Supervision, Project administration, Funding acquisition
\textbf{Philipp Tenbrock}: Conceptualization, Methodology, Software, Validation, Investigation, Data Curation, Writing - Original Draft, Writing - Review \& Editing, Visualization, Funding acquisition
\textbf{Balázs András Bálint}: Conceptualization, Methodology, Software, Validation, Investigation, Data Curation, Writing - Original Draft, Writing - Review \& Editing, Visualization
\textbf{David Traunecker}: Methodology, Formal analysis, Investigation, Data Curation
\textbf{Frank Nägele}: Writing - Review \& Editing, Supervision
\textbf{József Váncza}: Writing - Review \& Editing, Supervision
\textbf{Marco F. Huber}: Writing - Review \& Editing, Supervision

\section*{Declaration of competing interest}

The authors declare that they have no known competing financial interests or personal relationships that could have appeared to influence the work reported in this paper.

\section*{Acknowledgements}

The research presented in this paper has received funding from the Bundesministerium für Bildung und Forschung in the Rob-aKademI project (project number 01IS20009C).


\bibliographystyle{elsarticle-num}
\bibliography{91_references}

@inproceedings{Laemmle.2020c,
 author = {L{\"a}mmle, Arik and K{\"o}nig, Thomas and El-Shamouty, Mohamed and Huber, Marco F.},
 title = {Skill-based Programming of Force-controlled Assembly Tasks using Deep Reinforcement Learning},
 editor = {{Procedia CIRP}},
 booktitle = {Proceedings of the CIRP Conference on Manufacturing Systems (CMS)},
 year = {2020},
 doi = {10.1016/j.procir.2020.04.153}
}

@inproceedings{Laemmle.2022,
 author = {L{\"a}mmle, Arik and Xiang, Zheng and B{\'a}lint, Bal{\'a}zs Andr{\'a}s},
 title = {Extension of Established Modern Physics Simulation for the Training of Robotic Electrical Cabinet Assembly},
 pages = {1317--1322},
 volume = {107},
 editor = {{Procedia CIRP}},
 booktitle = {Proceedings of the CIRP Conference on Manufacturing Systems (CMS)},
 year = {2022},
 doi = {10.1016/j.procir.2022.05.151}
}

@inproceedings{Laemmle.2022c,
 author = {L{\"a}mmle, Arik and Tenbrock, Philipp and B{\'a}lint, B{\'a}lazs and N{\"a}gele, Frank and Kraus, Werner and V{\'a}ncza, J{\'o}zsef and Huber, Marco F.},
 title = {Simulation-based Learning of the Peg-in-Hole Process Using Robot-Skills},
 pages = {9340--9346},
 publisher = {IEEE},
 isbn = {978-1-6654-7927-1},
 booktitle = {Proceedings of the International Conference on Intelligent Robots and Systems (IROS)},
 year = {2022},
 doi = {10.1109/IROS47612.2022.9982212}
}

@inproceedings{Nagele.2018,
 author = {N{\"a}gele, Frank and Halt, Lorenz and Tenbrock, Philipp and Pott, Andreas},
 title = {A Prototype-Based Skill Model for Specifying Robotic Assembly Tasks},
 pages = {558--565},
 publisher = {IEEE},
 isbn = {978-1-5386-3081-5},
 booktitle = {Proceedings of the International Conference on Robotics and Automation (ICRA)},
 year = {2018},
 doi = {10.1109/ICRA.2018.8462885}
}

@inproceedings{Nagele.2019,
 author = {N{\"a}gele, Frank and Halt, Lorenz and Tenbrock, Philipp and Pott, Andreas},
 title = {Composition and Incremental Refinement of Skill Models for Robotic Assembly Tasks},
 pages = {177--182},
 publisher = {IEEE},
 isbn = {978-1-5386-9245-5},
 booktitle = {Proceedings of the International Conference on Robotic Computing (IRC)},
 year = {2019},
 doi = {10.1109/IRC.2019.00034}
}

@inproceedings{Inoue.2017,
  author    = {Inoue, Tadanobu and De Magistris, Giovanni and Munawar, Asim and Yokoya, Tsuyoshi and Tachibana, Ryuki},
  booktitle = {2017 IEEE/RSJ International Conference on Intelligent Robots and Systems (IROS)},
  title     = {Deep reinforcement learning for high precision assembly tasks},
  year      = {2017},
  volume    = {},
  number    = {},
  pages     = {819-825},
  doi       = {10.1109/IROS.2017.8202244}
}

@inproceedings{Todorov.2012,
  author    = {Todorov, Emanuel and Erez, Tom and Tassa, Yuval},
  booktitle = {2012 IEEE/RSJ International Conference on Intelligent Robots and Systems}, 
  title     = {MuJoCo: A physics engine for model-based control}, 
  year      = {2012},
  volume    = {},
  number    = {},
  pages     = {5026-5033},
  doi       = {10.1109/IROS.2012.6386109}
}

@inproceedings{Fan.2019,
  author    = {Fan, Yongxiang and Luo, Jieliang and Tomizuka, Masayoshi},
  booktitle = {2019 International Conference on Robotics and Automation (ICRA)},
  title     = {A Learning Framework for High Precision Industrial Assembly},
  year      = {2019},
  volume    = {},
  number    = {},
  pages     = {811-817},
  doi       = {10.1109/ICRA.2019.8793659},
  ISSN      = {2577-087X},
  month     = {May}
}

@article{DeSchutter.2007,
 author = {de Schutter, Joris and de Laet, Tinne and Rutgeerts, Johan and Decr{\'e}, Wilm and Smits, Ruben and Aertbeli{\"e}n, Erwin and Claes, Kasper and Bruyninckx, Herman},
 year = {2007},
 title = {Constraint-based Task Specification and Estimation for Sensor-Based Robot Systems in the Presence of Geometric Uncertainty},
 pages = {433--455},
 volume = {26},
 number = {5},
 journal = {International Journal of Robotics Research (IJRR)},
 doi = {10.1177/027836490707809107}
}

@article{Kakade.2020,
  title     = {Design Optimization of Snap Fit Feature of Lock Plate to Reduce Its Installation Force Using DoE Methodology},
  author    = {Kakade, Hrishikesh S and Sayyad, FB and Patil, Vinod G},
  journal   = {IUP Journal of Mechanical Engineering},
  volume    = {13},
  number    = {2/3},
  pages     = {83--96},
  year      = {2020},
  publisher = {IUP Publications},
  _url       = {https://www.academia.edu/download/60436434/IRJET-V6I660820190829-106062-173o44o.pdf}
}

@article{Amaya.2019,
  title     = {Detailed design process and assembly considerations for snap-fit joints using additive manufacturing},
  journal   = {Procedia CIRP},
  volume    = {84},
  pages     = {680-687},
  year      = {2019},
  note      = {29th CIRP Design Conference 2019, 08-10 May 2019, Póvoa de Varzim, Portgal},
  issn      = {2212-8271},
  doi       = {https://doi.org/10.1016/j                .procir.2019.04.271},   
  _url       = {https://www.sciencedirect.com/science/article/pii/S2212827119309187},
  author    = {Jorge Luis Amaya and Emilio A. Ramírez and Galarza F. Maldonado and Jorge Hurel},
  keywords  = {Design method, Additive manufacturing, Snap-fit},
  abstract  = {The use of additive manufacturing (AM) technology has been widely adopted due to the facility to produce highly complex elements compared to conventional fabrication processes. Additionally, AM technology is rapidly developing straightforward systems enabling designers to make products faster, despite current technology limitations (i.e. processing defects, materials properties, etc.). However, not only AM technology or products must be analyzed to have concrete solutions to all existing limitations. This means, it is necessary to take into account AM design process to propose simpler solutions. Elements manufactured by AM technology have dimension limitations on build size regarding printers building capabilities, especially when the elements are more volumetric than the building chamber. In those cases, AM design process takes a significant role and a potential solution is to divide big elements in sections, which are later 3D-printed and joined using snap-fits, as the cheapest and fastest connectors available. Thus, the present work explores the detail design stages of a proposed design methodology for elements´ coupling by snap-fit joints using AM technology. The design methodology is tested on the assembly of parts from a 1-gallon plastic container. A finite element simulation for the parts coupling scenarios is presented and the effects of part’s deflection on the detail design stages are analyzed. In addition, a final design validation regarding assembly ergonomics and retention forces are discussed in order to avoid part decoupling problems or material failure.}
}

@article{Kroemer.2019,
  author    = {Oliver Kroemer and
               Scott Niekum and
               George Dimitri Konidaris},
  title     = {A Review of Robot Learning for Manipulation: Challenges, Representations,
               and Algorithms},
  journal   = {CoRR},
  volume    = {abs/1907.03146},
  year      = {2019},
  _url       = {http://arxiv.org/abs/1907.03146},
  eprinttype = {arXiv},
  eprint    = {1907.03146},
  timestamp = {Tue, 19 Nov 2019 08:33:58 +0100},
  _biburl    = {https://dblp.org/rec/journals/corr/abs-1907-03146.bib},
  _bibsource = {dblp computer science bibliography, https://dblp.org}
}

@article{Ibarz.2021,
  author    = {Julian Ibarz and
               Jie Tan and
               Chelsea Finn and
               Mrinal Kalakrishnan and
               Peter Pastor and
               Sergey Levine},
  title     = {How to Train Your Robot with Deep Reinforcement Learning; Lessons
               We've Learned},
  journal   = {CoRR},
  volume    = {abs/2102.02915},
  year      = {2021},
  _url       = {https://arxiv.org/abs/2102.02915},
  archivePrefix = {arXiv},
  eprint    = {2102.02915},
  timestamp = {Tue, 09 Feb 2021 13:35:56 +0100},
  _biburl    = {https://dblp.org/rec/journals/corr/abs-2102-02915.bib},
  _bibsource = {dblp computer science bibliography, https://dblp.org},
  page      = {9}
}

@article{Tan.2018,
  author    = {Jie Tan and
               Tingnan Zhang and
               Erwin Coumans and
               Atil Iscen and
               Yunfei Bai and
               Danijar Hafner and
               Steven Bohez and
               Vincent Vanhoucke},
  title     = {Sim-to-Real: Learning Agile Locomotion For Quadruped Robots},
  journal   = {CoRR},
  volume    = {abs/1804.10332},
  year      = {2018},
  _url       = {http://arxiv.org/abs/1804.10332},
  archivePrefix = {arXiv},
  eprint    = {1804.10332},
  timestamp = {Mon, 13 Aug 2018 16:46:27 +0200},
  _biburl    = {https://dblp.org/rec/journals/corr/abs-1804-10332.bib},
  _bibsource = {dblp computer science bibliography, https://dblp.org}
}

@article{ElShamouty.2019,
  author    = {Mohamed El-Shamouty and
               Kilian Kleeberger and
               Arik Lämmle and
               Marco Huber},
  doi       = {doi:10.1515/teme-2019-0072 },
  _url       = {https://doi.org/10.1515/teme-2019-0072 },
  title     = {Simulation-driven machine learning for robotics and automation},
  journal   = {tm - Technisches Messen},
  number    = {11},
  volume    = {86},
  year      = {2019},
  pages     = {673--684}
}

@article{Kunz.2000,
 author = {Kunz, J},
 year = {2000},
 title = {Schnapphakenkr{\"a}fte mit neuem Ansatz genauer berechnen},
 number = {11},
 journal = {Kunststoffe-Synthetics}
}

@article{park.2020,
  author    = {Park, Hyeonjun and Park, Jaeheung and Lee, Dong-Hyuk and Park, Jae-Han and Bae, Ji-Hun},
  journal   = {IEEE Robotics and Automation Letters},
  title     = {Compliant Peg-in-Hole Assembly Using Partial Spiral Force Trajectory With Tilted Peg Posture},
  year      = {2020},
  volume    = {5},
  number    = {3},
  pages     = {4447-4454},
  doi       = {10.1109/LRA.2020.3000428},
  ISSN      = {2377-3766},
  month     = {July}
}

@article{Tang.2023,
    title={IndustReal: Transferring Contact-Rich Assembly Tasks from Simulation to Reality},
    journal={Robotics: Science and Systems (RSS)},
    author={Bingjie Tang and Michael A. Lin and Iretiayo Akinola and Ankur Handa and Gaurav S. Sukhatme and Fabio Ramos and Dieter Fox and Yashraj Narang},
    year={2023},
    eprint={2305.17110},
    archivePrefix={arXiv},
    primaryClass={cs.RO},
    url={https://arxiv.org/abs/2305.17110}, 
}

@article{Zhang.2023b,
year = {2023},
author = {Zhang, Xiang and Tomizuka, Masayoshi and Li, Hui}, 
title = {Bridging the Sim-to-Real Gap with Dynamic Compliance Tuning for  Industrial Insertion},
 url = {http://arxiv.org/pdf/2311.07499v1}
}

@article{Chen.2024,
author = {Chen, Chengjun and Zhang, Hao and Pan, Yong and Li, Dongnian},
year = {2024},
month = {01},
pages = {1-17},
title = {Robot autonomous grasping and assembly skill learning based on deep reinforcement learning},
volume = {130},
journal = {The International Journal of Advanced Manufacturing Technology},
doi = {10.1007/s00170-024-13004-0}
}

@article{Lin.2024,
      title={Generalize by Touching: Tactile Ensemble Skill Transfer for Robotic Furniture Assembly}, 
      author={Haohong Lin and Radu Corcodel and Ding Zhao},
      year={2024},
      eprint={2404.17684},
      archivePrefix={arXiv},
      primaryClass={cs.RO},
      url={https://arxiv.org/abs/2404.17684}, 
}

@article{Manyar.2026,
title = {Autonomous robotic screwdriving for high-mix manufacturing},
journal = {Robotics and Computer-Integrated Manufacturing},
volume = {98},
pages = {103172},
year = {2026},
issn = {0736-5845},
doi = {https://doi.org/10.1016/j.rcim.2025.103172},
url = {https://www.sciencedirect.com/science/article/pii/S0736584525002261},
author = {Omey M. Manyar and Rutvik Patel and Satyandra K. Gupta},
}

@article{Zhu.2026,
title = {Toward generalizable robotic assembly: A prior-guided deep reinforcement learning approach with multi-sensor information},
journal = {Robotics and Computer-Integrated Manufacturing},
volume = {100},
pages = {103242},
year = {2026},
issn = {0736-5845},
doi = {https://doi.org/10.1016/j.rcim.2026.103242},
url = {https://www.sciencedirect.com/science/article/pii/S0736584526000219},
author = {Zilu Zhu and Yongkui Liu and Qianji Wang and Zinan Wang and Lihui Wang and Sichao Liu and Bin Zi and Lin Zhang},
}

@article{Wang.2026,
title = {Digital twin-empowered robotic arm manipulation with reinforcement learning: A comprehensive survey},
journal = {Robotics and Computer-Integrated Manufacturing},
volume = {98},
pages = {103151},
year = {2026},
issn = {0736-5845},
doi = {https://doi.org/10.1016/j.rcim.2025.103151},
url = {https://www.sciencedirect.com/science/article/pii/S0736584525002054},
author = {Yichen Wang and Shuai Zheng and Ze Yang and Yingnan Zhu and Sen Zhang and Jiewu Leng and Jun Hong},
}

@article{Monnet.2025,
url = {https://doi.org/10.1515/auto-2024-0177},
title = {Leveraging reinforcement and curriculum learning for flexible robot-based snap-fit assembly automation},
author = {Josefine Monnet and Emma Heyen and Oliver Petrovic and Werner Herfs},
pages = {331--340},
volume = {73},
number = {5},
journal = {at - Automatisierungstechnik},
doi = {doi:10.1515/auto-2024-0177},
year = {2025},
lastchecked = {2026-02-12}
}

@phdthesis{Nagele.2021,
 author = {N{\"a}gele, Frank},
 year = {2021},
 title = {Prototypbasiertes Skill-Modell zur Programmierung von Robotern f{\"u}r kraftgeregelte Montageprozesse},
 school = {{Universit{\"a}t Stuttgart}},
 doi = {10.18419/OPUS-11655},
 type = {Dissertation}
}

@book{Rossdeutscher.2011,
 author = {Ro{\ss}deutscher, Mario},
 year = {2011},
 title = {Entwicklung eines Verfahrens zum Programmtest in der robotergest{\"u}tzten Montage: Zugl.: Cottbus, Techn. Univ., Diss., 2011},
 price = {EUR 48.80 (DE), EUR 48.80 (AT), sfr 97.60 (freier Pr.)},
 address = {Aachen},
 publisher = {Shaker},
 isbn = {9783844004649},
 series = {Berichte aus dem Lehrstuhl Automatisierungstechnik, BTU Cottbus}
}

@book{Gross.2017,
 author = {Gross, Dietmar and Hauger, Werner and Schr{\"o}der, J{\"o}rg and Wall, Wolfgang A.},
 year = {2017},
 title = {Technische Mechanik 2: Elastostatik},
 edition = {13. Aufl. 2017},
 publisher = {{Springer Berlin Heidelberg}},
 isbn = {9783662536797}
}

@inbook{classical.beam.models,
  author    = {Beer, Ferdinand P. and
               Johnston, E. Russel and
               Dewolf, John T. and
               Mazurek, David F.},
  title     = {Mechanics of Materials},
  edition   = {6},
  year      = {2012},
  publisher = {McGraw-Hill, a business unit of The McGraw-Hill Companies, Inc.},
  location  = {New York, NY, United States of America},
  isbn      = {978-0-07-338028},
  pages     = {A28}
}

@inbook{Troughton.2008,
  author    = {Michael J. Troughton},
  title     = {Handbook of plastics joining: a practical guide},
  edition   = {2},
  year      = {2008},
  publisher = {William Andrew, Inc.},
  address   = {Norwich, NY, United States of America},
  _url       = {https://books.google.de/books?id\=BXL\_mnDzW0QC\&pg\=PA188},
  chapter   = {18.6}
}

@techreport{Tempel.2017,
  author    = {Tempel, Philipp and
               Eger, Florian and
               Lechler, Armin and
               Verl, Alexander},
  title     = {Control Cabinet Manufacturing 4.0: A study on the potential of automation and digitalisation in the manufacturing of control cabinets and switchgears in machine and systems engineering},
  institution = {EPLAN Software \& Service GmbH \& Co. KG},
  year      = {2017},
  address   = {Monheim am Rhein, Germany},
  _url       = {https://www.eplan-software.com/control-cabinet-engineering-40-study/}
}

@misc{FEM.creo,
  organization = {PTC Inc.},
  address   = {Boston, MA, United States of America},
  title     = {Creo Ansys Simulation: Engineering Simulation Software},
  year      = {2021},
  month     = {8},
  url       = {https://www.ptc.com/en/products/creo/ansys-simulation}
}

@misc{FEM.solidworks,
  organization = {Dassault Systèmes SolidWorks Corporation},
  address   = {Waltham, MA, United States of America},
  title     = {SOLIDWORKS Simulation},
  year      = {2021},
  month     = {8},
  url       = {https://www.solidworks.com/product/solidworks-simulation}
}

@misc{FEM.inventor,
  organization = {Autodesk Inc.},
  address   = {San Rafael, CA, United States of America},
  title     = {Inventor Features: 2022, 2021 Features},
  year      = {2021},
  month     = {8},
  url       = {https://www.autodesk.com/products/inventor/features}
}

@online{StabelBaselines3,
  title     = {Stable-Baselines3 Docs - Reliable Reinforcement Learning Implementations},
  url       = {https://stable-baselines3.readthedocs.io/en/master/},
  urldate = {10.01.2026}
}

\clearpage\onecolumn

\end{document}